\documentclass[runningheads]{llncs}
\usepackage{graphicx}
\usepackage{amsmath,amssymb} 
\usepackage{color}
\usepackage{figsize}
\usepackage{multirow}
\usepackage{bbm}
\usepackage{algorithm}
\usepackage{algorithmic}

\begin{document}

\title{TextNet: Irregular Text Reading from Images\\ with an End-to-End Trainable Network}

\titlerunning{TextNet} 

\author{Yipeng Sun$^\star$ \and Chengquan Zhang$^\star$ \and Zuming Huang \and Jiaming Liu \and Junyu Han \and Errui Ding}


%

\authorrunning{Y. Sun, C. Zhang et al.} 


\institute{Baidu Inc, Beijing, China.\\
\email{\{sunyipeng,zhangchengquan,huangzuming,liujiaming03,hanjunyu,dingerrui\}@baidu.com}}

\maketitle

\begin{abstract}
Reading text from images remains challenging due to multi-orientation, perspective distortion and especially the curved nature of irregular text. Most of existing approaches attempt to solve the problem in two or multiple stages, which is considered to be the bottleneck to optimize the overall performance. To address this issue, we propose an end-to-end trainable network architecture, named~\textit{TextNet}, which is able to simultaneously localize and recognize irregular text from images. Specifically, we develop a scale-aware attention mechanism to learn multi-scale image features as a backbone network, sharing fully convolutional features and computation for localization and recognition. In text detection branch, we directly generate text proposals in quadrangles, covering oriented, perspective and curved text regions. To preserve text features for recognition, we introduce a perspective RoI transform layer, which can align quadrangle proposals into small feature maps. Furthermore, in order to extract effective features for recognition, we propose to encode the aligned RoI features by RNN into context information, combining spatial attention mechanism to generate text sequences. This overall pipeline is capable of handling both regular and irregular cases. Finally, text localization and recognition tasks can be jointly trained in an end-to-end fashion with designed multi-task loss. Experiments on standard benchmarks show that the proposed \textit{TextNet} can achieve state-of-the-art performance, and outperform existing approaches on irregular datasets by a large margin.
\end{abstract}
\vspace{-1.5em}


\section{Introduction}
\vspace{-0.5em}
Reading text from images is one of the most classical and elemental problems in pattern recognition and machine intelligence. This problem has received much attention due to its profound impact and valuable applications in both research and industrial communities. End-to-end text reading is to tell the locations and content of text from images, combining text detection and recognition. Benefiting from the deep learning paradigm for generic object detection \cite{girshick2015fast}\cite{ren2015faster}\cite{huang2015densebox}\cite{liu2016ssd} and sequence-to-sequence recognition \cite{graves2006connectionist}\cite{bahdanau2014neural}\cite{xu2015show}, recent advances have witnessed dramatic improvement in terms of both recognition accuracy and model simplicity for text localization \cite{2017wordsup}\cite{2017east}\cite{2017deep_matching} and recognition~\cite{shi2017end}\cite{shi2016robust}\cite{lee2016recursive}. Regarded as the ultimate goal of text reading, end-to-end task can be tackled by integrating text detection and recognition algorithms into an end-to-end framework. 
\begin{figure}
\centering
\SetFigLayout{2}{1}
  \subfigure
  {\includegraphics[width=0.35\textwidth, height=0.2\textwidth]{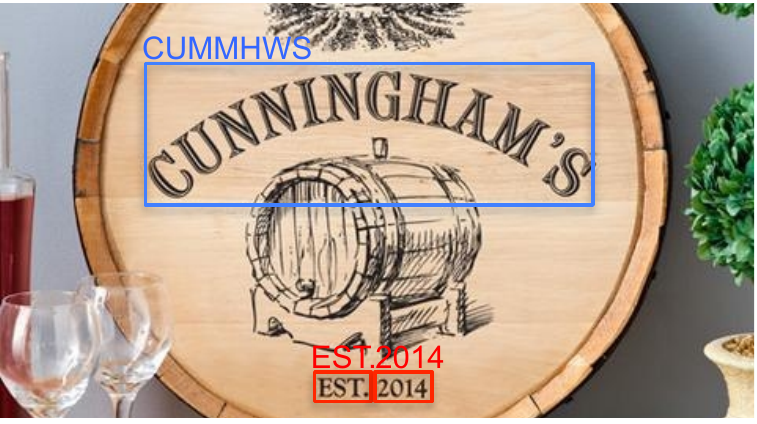}}
  \subfigure
   {\includegraphics[width=0.35\textwidth, height=0.2\textwidth]{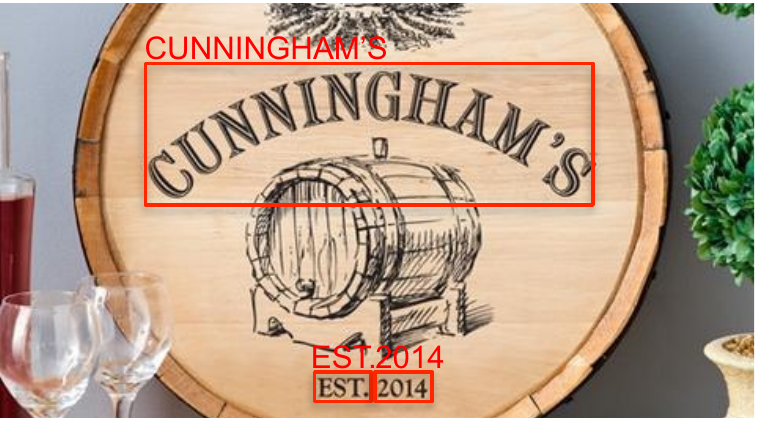}}
\vspace{-1em}
\caption{\footnotesize{End-to-end results of irregular text: (a) existing approaches (b) the proposed TextNet results.}}
\label{fig:curved_text}
\vspace{-1.5em}
\end{figure}

Conventional end-to-end text reading paradigm typically works as a two-stage or multi-stage system, integrating detection and recognition modules into an overall pipeline~\cite{wang2012end}\cite{jaderberg2016reading}. Recent text detection approaches are able to generate text locations in an end-to-end trained model~\cite{2017east}\cite{2017deep_matching}. The state-of-the-art text recognition methods can be formulated as an end-to-end "translation" problem from a cropped text image to text sequence~\cite{shi2017end}\cite{lee2016recursive}. More recently, in contrast to multi-stage text reading systems, fully trainable models to tackle end-to-end text reading task have been proposed in~\cite{2017towards}\cite{2017deep_textspotter}\cite{2018textspotter}. These approaches are capable of predicting locations and content of regular text in one model. However, these state-of-the-art approaches face difficulties in handling irregular text with perspective distortion, and fail to tackle curved text cases. Therefore, it is still challenging to localize and recognize irregular text in an end-to-end framework, and it is an open issue to read both regular and irregular text in a unified model.

Towards this end, we propose an end-to-end trainable text reading network, named \textit{TextNet}, which is able to read irregular text from images, especially for perspective and curved cases. Taking a full image as the input, the proposed network can generate word proposals by directly regressing quadrangles to cover both oriented and curved text regions. To preserve text features for recognition, we develop a perspective RoI transform layer to align these proposals into small feature maps. To extract more effective features from text proposals, we propose to encode the aligned RoI~(Region-of-Interest) features in RNN~(Recurrent Neural Network), and utilize the spatial attention mechanism to recognize both regular and irregular text sequences. Text localization and recognition tasks are designed as an end-to-end trainable network by multi-task learning. The contributions of this paper are listed as follows:\vspace{-0.5em}
\begin{itemize}
\item We propose an end-to-end trainable network that can simultaneously localize and recognize irregular text in one model.
\item To cover perspective and curved text regions, we directly predict quadrangle text proposals, and develop a perspective RoI transform layer to align these proposals into small feature maps, keeping invariance in aspect ratio.
\item To extract effective text features for recognition, we propose to encode the aligned features using RNN in both vertical and horizontal directions, and utilize spatial attention mechanism to decode each character despite of imprecise predicted locations.
\item Experiments on datasets demonstrate that the proposed~\textit{TextNet} can achieve comparable or state-of-the-art performance on the standard benchmarks, and outperform the existing approaches on irregular datasets by a large margin.
\end{itemize}
To our best knowledge, it is the first end-to-end trainable text reading network, which is able to handle oriented, perspective and curved text in a unified model.

\vspace{-1em}
\section{Related Work}
\vspace{-0.5em}
In this section, we will summarize the existing literatures on text localization and text recognition, respectively. As the combination of both text localization and recognition, recent advances in end-to-end text reading and spotting will also be discussed.

\textbf{Text localization} aims to tell the locations of words or text-lines from images. Conventional approaches can be roughly classified as components-based~\cite{neumann12mser}\cite{tian2016ctpn} and character-based methods~\cite{2017wordsup}. These methods first attempt to find local elements, such as components, windows, or characters, and then group them into words to generate the final bounding boxes. In recent years, most of these approaches utilize deep convolutional neural network to learn text features, which has shown significant performance improvement. Inspired by recent advances in generic object detection~\cite{ren2015faster}\cite{huang2015densebox}\cite{liu2016ssd}, the state-of-the-art text detection approaches~\cite{2017east}\cite{2017deep_matching} are designed to directly predict word-level bounding boxes, which simplifies the previous multi-stage pipeline into an end-to-end trainable model. These approaches can be further classified as proposed-based or single-shot models. To tackle multi-scale and orientation problems in text detection, Liao et al.~\cite{liao2017textboxes} employ SSD framework with specifically designed anchor boxes to cover text regions. Liu et al.~\cite{2017deep_matching} develop a rotational proposal network to localize text rotational bounding boxes. Following the general design of DenseBox~\cite{huang2015densebox}, Zhou et al.~\cite{2017east} and He et al.~\cite{he2017direct} propose to directly regress multi-oriented text in quadrangle representation. 

\textbf{Text recognition} aims to assign text labels to each character, taking a cropped word or a text-line as input. Traditional text recognition methods are generally considered as character-based~\cite{bissacco2013photoocr}, or word-based classification~\cite{gupta2016synthetic} problems. These paradigms are either composed of multiple steps or difficult to generalize to non-Latin script languages. In recent years, the success of recurrent neural network has inspired the text recognition approaches to be formulated as a variable-length sequence-to-sequence problem. Following the paradigms of speech recognition~\cite{graves2013speech} and handwritten recognition, Pan et al.~\cite{he2016reading} and Shi et al.~\cite{shi2017end} propose to extract convolutional features, reshape to an one-dimensional sequence and encode context in RNN for recognition. The whole model can be trained as a sequence labeling problem with CTC~(Connectionist Temporal Classification)~\cite{graves2006connectionist} loss. These approaches make it possible to train word-level and line-level text recognition in an end-to-end trainable framework. Most recently, with the neural machine translation break-through by attention mechanism~\cite{bahdanau2014neural}, several attention-based text recognition models~\cite{shi2016robust}\cite{lee2016recursive} are developed to encode convolutional image features into sequence context with RNN, and predict each character using attention-based decoder. 

\textbf{End-to-end text reading} is typically recognized as the ultimate evaluation of text recognition from images. Building upon both text detection and recognition algorithms, traditional approaches are mostly composed of multiple steps to achieve end-to-end results~\cite{wang2012end}\cite{jaderberg2016reading}\cite{liao2018textboxes++}. Jaderberg et al.~\cite{jaderberg2016reading} first generate text proposals in high recall, and then refine the bounding boxes to estimate more precisely. Finally, cropped word images are recognized by a CNN-based word classifier. In contrast to traditional methods in multiple steps, recently, there are a number of approaches designed in a fully end-to-end trainable framework. Li et al.~\cite{2017towards} propose an end-to-end trainable text reading model, using RPN~(Region Proposal Network) to estimate proposals and LSTM with attention to decode each words. To address the multi-orientation problem of text, Busta et al.~\cite{2017deep_textspotter} utilize YOLOv2~\cite{redmon2016yolo9000} framework to generate rotational proposals, and train RoI sampled features with CTC loss. Other approaches are designed to further speed up computation for inference~\cite{2018fots}. All these approaches are trainable in an end-to-end fashion, which can improve the overall performance and save computation overhead by alleviating multiple processing steps. 
However, it is still challenging to handle both horizontal, oriented and perspective, especially curved text in an end-to-end framework. As one of the ultimate goals for end-to-end text reading, this problem has not been investigated yet in the existing literatures. 

\vspace{-1em}
\section{Model Architecture}
\begin{figure}
\begin{center}
\includegraphics[width=0.78\textwidth]{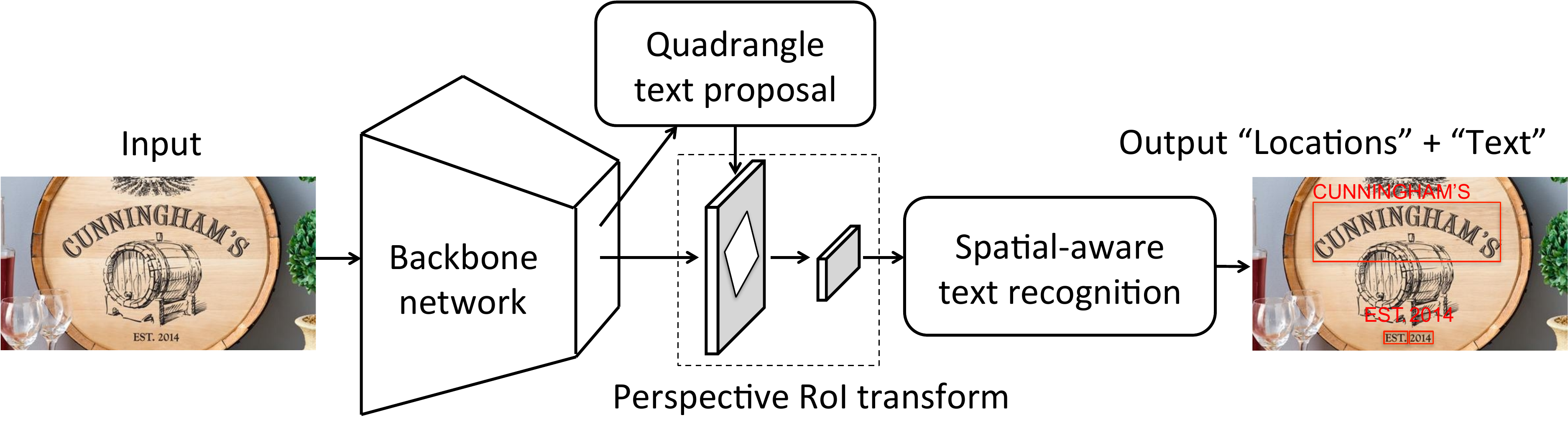}
\vspace{-1em}
\caption{\footnotesize{The overall architecture of \textit{TextNet}.}}
\label{fig:framework}
\end{center}
\vspace{-2em}
\end{figure}

\vspace{-0.5em}
TextNet is an end-to-end trained network capable of localizing and recognizing both regular and irregular text simultaneously. As shown in Fig.~\ref{fig:framework}, the overall network architecture consists of four building blocks, i.e., the backbone network, quadrangle text proposal, perspective RoI transform and spatial-aware text recognition. 
To enable the joint training of text detection and recognition tasks, we derive a backbone network to share deep convolution features, and aggregate the multi-scale features into $\frac{1}{4}$ resolution of the input image. In our work, we utilize ResNet-50~\cite{he2016deep} as the backbone network for its ease-of-use. To learn multi-scale text features, we develop a scale-aware attention mechanism to fuse multi-scale features to generate the final feature map. To cover oriented, perspective and curved text regions, text proposal network can directly predict the locations of text in quadrangles. Furthermore, in order to preserve information from quadrangle proposals for recognition, we develop a perspective RoI transform layer to convert the features of quadrangle proposals into fixed-height features with variable-size width. Finally, as the recognition network, we propose to use RNN as feature context encoder and combine spatial attention-based RNN decoder, thus improving the recognition performance despite of imperfect detection as well as perspective and curved text distortion. The overall architecture containing both detection and recognition branches can be jointly trained in an end-to-end manner. In this section, we will describe each part of the model in detail, respectively.

\vspace{-1em}
\subsection{Backbone Network}
\vspace{-0.5em}
Using ResNet-50 as a shared feature extractor, we manage to make use of features in $4$-scales to generate the final feature map. Since visual objects usually vary in size, it becomes difficult to detect multi-scale objects. In recent deep learning paradigm, it has shown its effectiveness to aggregate multi-scale features to localize objects. Existing literatures have introduced a U-shape network~\cite{ronneberger2015u} to merge multi-scale features, build a feature pyramid to boost the object detection performance~\cite{lin2017feature}, and fuse multi-scale convolutional features by attention mechanism to improve image segmentation accuracy~\cite{chen2016attention}. Since the scales of text usually vary in much a wider range than generic objects, feature maps from low to high resolutions have different responses corresponding to the scale of text. Therefore, there exists an optimal feature map scale for each text to maximize its response.

\begin{figure}
\begin{center}
\includegraphics[width=0.8\textwidth]{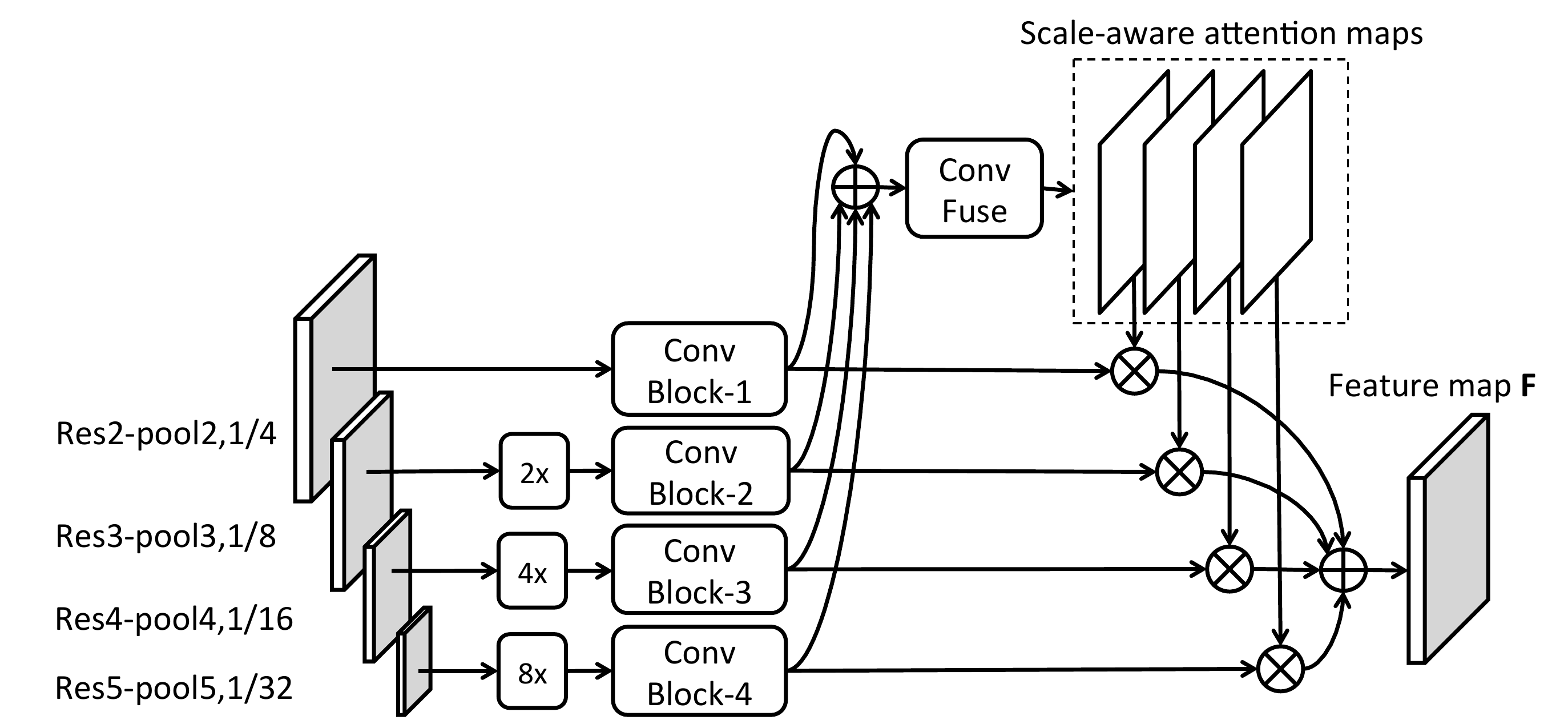}
\vspace{-1em}
\caption{\footnotesize Scale-aware attention mechanism for multi-scale features aggregation.}
\label{fig:scale-attention}
\end{center}
\vspace{-2.5em}
\end{figure}

Towards this end, inspired by the multi-scale attention mechanism in image segmentation~\cite{chen2016attention}, we come up with the idea of learning to fuse features by adaptively weighting multi-scale feature maps. In our work, we develop a scale-aware attention mechanism to fuse features at different scales, adaptively merging them into one feature map. This can help to improve the recall and precision of both large and small characters. As shown in  Fig.~\ref{fig:scale-attention}, the developed backbone network takes four scales of features from ResNet-50 as its input, transform these features into the same scale at $1/4$ by upsampling, and predict attention maps corresponding to features at different scales. The final feature map is the element-wise addition of $4$-scale features, which are spatially weighted by attention mechanism. In this network, $\textrm{Conv-Block-1, 2, 3, 4}$ are standard convolutional operations of $($conv $3\times3$, $256)$ + $($conv $1\times1$, $128)$, and Conv-Fuse is a feature fusion block by $($conv $1\times1$, $4)$ operation, generating attention weights in $4$ scales. With the developed structure, the backbone network can learn multi-scale features of text, thus improving the detection and recognition performance. 

\vspace{-1em}
\subsection{Quadrangle Text Proposal}
\vspace{-0.5em}
To improve the recall of text regions, we utilize the quadrangle representation to define a text proposal, following the design of EAST~\cite{2017east}. The text detection branch can directly generate a score map~$\mathbf{S}=\{p_{ij} | i,j \in \Omega \}$, where $\Omega$ is defined as the feature region. The score map is used to distinguish between text and non-text with probability $p_{ij}$ as being text. Besides, $4$-point coordinate offsets are estimated for every position to form quadrangle representation, which can be defined as $\mathbf{Q} = \{(\Delta x_k,\Delta y_k)_{k=1,2,3,4}\}$. Taking the finally aggregated feature map~$\mathbf{F}$ as input, the output channels in text detection branch are predicted in fully convolutional operations by $\mathbf{Q}=\mathrm{conv}_{3 \times 3, 8}(\mathbf{F})$ and $\mathbf{S}=\mathrm{conv}_{3 \times 3, 1}(\mathbf{F})$.
Finally, the text proposals in quadrangles are calculated by NMS~(Non-Maximum Suppression) in the predicting stage, following state-of-the-art object detection pipeline.

During the training stage, the text detection loss $L_{det}$ is composed of two parts, which can be defined as $L_{det} = L_{quad} + \lambda L_{cls}$. In this equation, $L_{cls}$ measures the difference between the predicted and ground-truth score maps by cross-entropy loss. $L_{quad}$ is defined as the smooth $\textrm{L}_{1}$ loss to measure the distance between the predicted coordinates and ground-truth locations, and $\lambda$ is the regularization parameter that can control the trade-off between two loss functions.

\vspace{-1em}
\subsection{Perspective RoI Transform}
\vspace{-0.5em}
The conventional RoI-pooling \cite{girshick2015fast} extracts a small fixed-size feature map from each RoI (e.g., $8\times8$). To improve image segmentation accuracy, RoI-Align introduced by Mask R-CNN~\cite{he2017mask} aims to solve the misalignment between input and output features. To better design for text regions, the varying-size RoI pooling \cite{2017towards} has been developed to keep the aspect ratio unchanged for text recognition. Moreover, to handle oriented text regions, rotational text proposal~\cite{2017deep_textspotter} and RoI-Rotate~\cite{2018fots} have been developed to address this issue using affine transformation, motivated by the idea of Spatial Transformer Network~\cite{jaderberg2015spatial} to learn a transformation matrix.

By contrast, following the generated quadrangle proposals, we develop perspective RoI transform to convert an arbitrary-size quadrangle into a small variable-width and fixed-height feature map, which can be regarded as the generalization of the existing methods. Our proposed RoI transform can warp each RoI by perspective transformation and bilinear sampling. The perspective transformation $T_\theta$ can be calculated between the coordinates of a text proposal and the transformed width and height. Each RoI from the feature map can be transformed to axis-aligned feature maps by perspective transformation. The perspective transformation is denoted as $T_\theta =  [\theta_{11}, \theta_{12}, \theta_{13}; \theta_{21}, \theta_{22}, \theta_{23}; \theta_{31}, \theta_{32}, 1]$,
and $w^t$, $h^t$ denote the width and height of a transformed feature map, respectively.
The parameters of $T_\theta$ can be calculated using the source and destination coordinates, following the principle of perspective transformation.
Using these parameters, the transformed feature map can be obtained by perspective transformation as \vspace{-1em}
\begin{equation}\label{eq:perspective_transformation}
\begin{split}
\begin{pmatrix}
u\\
v\\
w
\end{pmatrix} &= 
T_{\theta}
\begin{pmatrix}
x^t_k\\
y^t_k\\
1
\end{pmatrix},\vspace{-0.5em}
\end{split}
\end{equation}
where $(x^t_k, y^t_k)$ are the target coordinates from the transformed feature map with pixel index $k$ for $\forall k = 1, 2, ..., h^tw^t$, and $u$, $v$, and $w$ are auxiliary variables. The source coordinates $(x^s_k, y^s_k)$ from the input are defined as $x^s_k = u / w$ and $y^s_k = v / w$, respectively. The pixel value of $(x^t_k, y^t_k)$ can be computed by bilinear sampling from the input feature map as $V_k = \sum^{h^s}_n \sum^{w^s}_m U_{nm} K(x^s_k - m) K(y^s_k - n)$, where $V_k$ is the output value of pixel $k$, $h^s$ and $w^s$ are the height and width of the input. In this equation, $U_{nm}$ denotes the value at location $(n, m)$ from the input, and the kernel function is defined as $K(\cdot) = \max(0, 1 - |\cdot|)$. Note that the bilinear sampling operates on each channel, respectively.

\vspace{-1em}
\subsection{Spatial-aware Text Recognition}
To handle irregular cases, the proposed text detection network attempts to cover text regions by quadrangle proposals, improving the performance in terms of recall. In an end-to-end text reading task, it is crucial to design a concatenation module between text detection and recognition to achieve better accuracy. The detected text regions may not be precise enough for text recognition approaches, especially in irregular text cases. The conventional text recognition approaches~\cite{shi2017end}\cite{lee2016recursive} assume that the input words or text lines have been well cropped and aligned. Therefore, in perspective and curved text cases, it is difficult to capture effective features in these approaches to recognize characters, which leads to failure cases. Recent studies have paid attention to irregular text recognition problem~\cite{shi2016robust}\cite{yang2017learning}. These approaches utilize cropped words provided by the datasets. From the perspective of end-to-end task, however, it has not been investigated yet for irregular text recognition. The spatial attention mechanism has been applied in road name recognition~\cite{2017attention} and handwritten recognition~\cite{bluche2016scan}\cite{bluche2016joint}, however, it is still limited to specific applications and has not shown effective yet for general scene text recognition. 

\begin{figure}
\begin{center}
\includegraphics[width=0.65\textwidth]{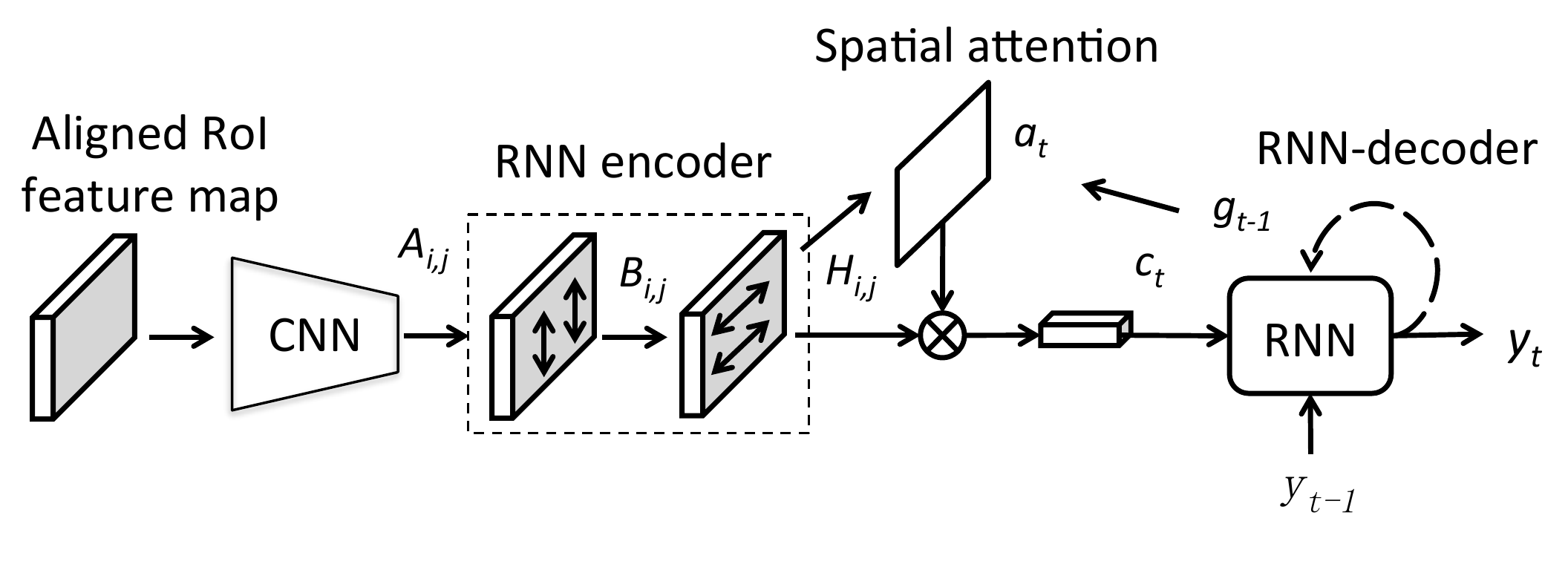}
\vspace{-1.5em}
\caption{\footnotesize Spatial-aware text recognition network}
\label{fig:spatial_att}
\end{center}
\vspace{-2.5em}
\end{figure}

Taking aligned RoI features as input, we propose to encode convolutional features by RNN to extract spatial context, and combine spatial attention mechanism~\cite{xu2015show}, forming an encoder-decoder architecture with attention mechanism to sequentially decode each character. This approach aims to attend to features-of-interest to cover both horizontal, oriented, perspective, and curved cases, extracting effective character features at each time step. As illustrated in Fig.~\ref{fig:spatial_att}, the proposed text recognition network takes each RoI aligned feature map as the input, and predicts each character label $y_t$ as the output. 

\textbf{Encoder} The encoder part consists of stacked convolutional operations and two layers of RNN. The convolutional operations are $4$-layer of (conv $3\times3$, 128) followed by batch normalization and ReLU. The first layer of RNN encodes each columns of the feature map $A \in R^{h_{A}\times w_{A} \times c_{A}}$ as\vspace{-0.5em}
\begin{equation}\label{eq:rnn_column}
h^{c}_{i} = RNN(A_{i,j}, h^{c}_{i-1}), \forall j = 1,2,\cdots,w_{A},
\end{equation}
where  $h^{c}_{i}$ is the hidden state in the $i$-th row. The second layer of RNN encoder takes the output $B \in R^{h_B \times w_B \times c_B}$ of the first layer, and encodes each row of the feature map as\vspace{-0.5em}
\begin{equation}
h^{r}_{j} = RNN(B_{i,j}, h^{r}_{j-1}), \forall i = 1,2,\cdots,h_{B}, 
\end{equation}
where $h^{r}_{j}$ is the hidden state in the $j$-th column. In our work, we use GRU~(Gated Recurrent Unit) as RNN encoder for its ease-of-use. After two layers of RNN operations, the final generated feature map $H \in R^{h_{H}\times w_{H} \times c_{H}}$ contains context information in both horizontal and vertical directions.


\textbf{Decoder} In the decoder part, spatial attention is to calculate the similarity between encoder feature map $H_{i,j}$ and RNN decoder state $g_{t-1}$, which is to learn the spatial alignment \vspace{-0.5em}
\begin{equation}\label{eq:attention}
e_{i,j,t} = Attention(H_{i,j}, g_{t-1})
\end{equation}
to weight the importance of feature $H_{i,j}$ at time $t$. In Eq.~\ref{eq:attention}, we utilize a feed forward neural network as $Attention(\cdot)$ function following~\cite{bahdanau2014neural}, and normalize the alignment as\vspace{-0.5em}
\begin{equation}\label{eq:normalize}
\alpha_{i,j,t} = \frac{\mathrm{exp}(e_{i,j,t})}{\sum_{i,j} \mathrm{exp}(e_{i,j,t})}.
\end{equation}
With the predicted spatial attention weights $\alpha_{i,j,t}$, the context vector $c_t$ at time $t$ is calculated as\vspace{-0.5em}
\begin{equation}\label{equ:spatial_att}
c_t = \sum_{i,j,t} \alpha_{i,j,t} H_{i,j}.
\end{equation}
Feeding $c_t$ and the former output $y_{t-1}$ as input, the RNN decoder can directly update its hidden state $g_t$ and predict the output label $y_t$ by fully connected weights and softmax classification. The final output is to achieve the character that can maximize the posterior probability as $\hat{y}_{t} = \textrm{arg}\max_{y_t \in \mathcal{D}} p(y_t|g_t, y_{t-1})$, where $\mathcal{D}$ is the set of labels. During the training stage, the recognition loss for sequences is derived as the average uncertainty, which can be defined as 
\begin{equation}
L_{reg} = - \frac{1}{NT}\sum^{N}_{n=1} \sum^{T}_{t=1}\log p(y_t|y_{t-1}, g_{t-1}, c_t). 
\end{equation}
Training this objective function is to minimize $L_{reg}$ over all training samples and time steps. In this equation, $N$ and $T$ denote the number of training samples and the count of time steps, respectively.

\vspace{-1em}
\subsection{Joint Localization and Recognition}
\vspace{-0.5em}
Benefiting from the developed modules above, we are able to unify the quadrangle text detection, perspective RoI transform, and spatial-aware text recognition network in one model. To train localization and recognition networks simultaneously, we utilize a multi-task loss defined as\vspace{-0.5em}
\begin{equation} \label{equ:multi_task_loss}
L = L_{det} + \beta L_{reg}, 
\end{equation}
where $\beta$ is the regularization parameter that can balance between two network branches. During the joint training stage, the input of the model consists of image samples, ground-truth labels of locations and the corresponding text. Text locations in quadrangles can be converted to geometry and score maps $\mathbf{G} = \{\mathbf{Q}_n, \mathbf{S}_n\}_{n=1,...,N}$. With the provided table of characters, the ground-truth text labels can be mapped to the corresponding class labels. During training and testing, the \textit{START} and \textit{EOS}~(End-of-Sequence) symbols are added to tell the RNN decoder when to start and end in text sequence recognition. 

To jointly train text localization and recognition network, the overall training procedure is divided into two stages. To overcome the shortage of small datasets, we utilize the vgg-synthetic data~\cite{gupta2016synthetic}, i.e., VGG-Synth 800k, to train the base network using ImageNet pretrained model. Specifically, one training strategy is to train the detection branch until it almost converges to a steady point, and jointly train detection and recognition branches simultaneously. Another training strategy is to train the recognition branch at first instead, and then jointly train the whole network. These two training procedures are equal to achieve the final convergence in our attempts. To evaluate the performance on benchmarks, we finetune the VGG-Synth trained model on ICDAR-13, ICDAR-15, and Total-Text, respectively. To ease training and improve generalization by data argumentation, we randomly crop samples from images, resize the long-side to $512$ pixels with mean value paddings. In experiments, we use multi-GPU clusters to train the model. The batch-size is $16$ per GPU, and the number of RoIs in a batch is set to $32$ per GPU for the recognition branch. The optimization algorithm is Adam, and the initial learning rate is set to $1\times10^{-4}$.

\vspace{-1em}
\section{Experiments}
\vspace{-1em}
To validate the performance of the proposed model, we conduct experiments on standard benchmarks, e.g., ICDAR-13, ICDAR-15,  as the regular datasets, and the curved dataset, e.g., Total-Text, as the irregular dataset. We evaluate and compare the performance of the proposed model on end-to-end and word spotting tasks with other competitive approaches.

\vspace{-1em}
\subsection{Datasets}
\subsubsection{ICDAR-13 and ICDAR-15} The widely used benchmarks for text reading include ICDAR-13 and ICDAR-15~\cite{karatzas15icdar} datasets. These datasets come from ICDAR 2013 and ICDAR 2015 Robust Reading Competitions, respectively. ICDAR-13 mainly includes horizontal text as focused scene text with word-level bounding box annotations and text labels. There are $229$ images for training and $233$ images for testing. Images from the ICDAR-15 dataset  are captured in an incidental way with Google Glass, and word-level quadrangle annotations and text labels are available. There are $1000$  images for training and $500$ for testing. \vspace{-1.5em}
\subsubsection{VGG synth} We also utilize the VGG synthetic dataset~\cite{gupta2016synthetic}, which consists of $800,000$ images. Synthetic text strings are well rendered and blended with background images. The dataset provides detailed character-level, word-level and line-level annotations, which can be used for model pre-training.\vspace{-1.5em}
\subsubsection{Total-Text} The Total-Text dataset~\cite{chng17tt} released in ICDAR 2017 is a collection of irregular text. Unlike the previous ICDAR datasets, there are a number of curved text cases with multiple orientations. There are $1255$ images in the training set, and $300$ images in the test set with word-level polygon and text annotations. 

\vspace{-1em}
\subsection{Quantitative Comparisons with Separate Models} 
\vspace{-0.5em}
To validate the effectiveness of end-to-end training, we compare the text detection results of  TextNet with the proposed detection branch only as a baseline. The evaluation protocols of text detection exactly follow the public criterion of ICDAR competitions, including the ICDAR-13, and ICDAR-13 DetEval and ICDAR-15. The IoU~(intersection-of-union) threshold is $0.5$ as the default value to decide whether it is a true positive sample or not. As shown in Tab.\ref{tab:icdar-13-det}, our proposed TextNet achieves $91.28\%$, $91.35\%$ and $87.37\%$ in F-measure under ICDAR-13, ICDAR-13 DetEval and ICDAR-15 criterion, respectively. The end-to-end trained model can achieve $+6.66\%$, $+6.65\%$ and $+4.13\%$ absolute F-measure improvement, compared with the detection model without recognition branch. Different from conventional text detectors, the joint trainable model with text recognition branch can help improve the representation power using shared backbone features, and improve both recall and precision during the detection stage. 

For irregular cases, we conduct ablation experiments on Total-Text dataset, and compare the end-to-end trained results with the two-stage approach, which is developed using separately trained detection and recognition models. As shown in Tab.~\ref{tab:total-text-det}, the proposed TextNet can obtain substantial improvement over the two-stage approach in detection and end-to-end tasks in terms of F-measure. Different from previous two-stage and end-to-end training methods, the proposed approach can tackle irregular cases in a unified framework, especially curved text, which further demonstrate the effectiveness of TexNet.

\vspace{-0.5em}
\subsection{Quantitative Comparisons with State-of-the-Art Results}
\subsubsection{Detection results} To validate the text detection performance, we compare the proposed TextNet with other state-of-the-art text detection methods.  Quantitative results on ICDAR-13 and ICDAR-15 in terms of recall, precision and F-measure are listed in Tab.\ref{tab:icdar-13-det}, respectively.  From the table, we find that the proposed algorithm can achieve state-of-the art results with at least $+3\%$ improvement in F-measure on ICDAR-13 and ICDAR-15 datasets. Note that * indicates the corresponding results have not been released in the original paper. The proposed jointly trainable model can improve the representation and generalization of text features in a shared backbone network by modeling detection and recognition in a unified network. To tackle regular and irregular cases with a two-directional RNN encoder, the spatial attention-based recognition branch can not only improve detection precision compared to other competitive results, but also increase the detection accuracy in terms of recall. 
\vspace{-1em}
\begin{table}
\begin{center}
\footnotesize 
\caption{\footnotesize Text detection results on ICDAR-13 and ICDAR-15 for comparisons. Note that R, P and F in this table are short for recall, precision, and F-measure, respectively. In the table, `TextNet-detection only' indicates that the model is trained for text detection without the help of recognition branch, and `TextNet' is the end-to-end trainable model to localize and recognize results simultaneously.}
\label{tab:icdar-13-det}
\begin{tabular}{|c||c|c|c||c|c|c||c|c|c|}
  \hline
  \multirow{2}{*}{Method} &
  \multicolumn{3}{c||}{ICDAR-13} &
  \multicolumn{3}{c||}{ICDAR-13 DetEval} & 
  \multicolumn{3}{c|}{ICDAR-15}\\ 
  \cline{2-10} 					 	& R & P & F & R & P & F & R & P & F \\
  \hline\hline
  TextBoxes~\cite{liao2017textboxes} 	& 83 & 88 & 85 & 89 & 83 & 86 & * & * & *  \\
  \hline
  CTPN~\cite{tian2016ctpn} 			& * & * & * & 83 & 93 & 88 & 52 & 74 & 61 \\
  \hline
  Liu et al.~\cite{2017deep_matching}        & * & * & * & * & * & * & 68.22 & 73.23 & 70.64 \\
  \hline
  SegLink~\cite{shi2017detecting} 		& * & * & * & 83.0 & 87.7 & 85.3 & 76.8 & 73.1 & 75.0\\
  \hline
  SSTD~\cite{2017deep_textspotter} 		& 86 & 88 & 87 & 86 & 89 & 88 & 73.86 & 80.23 & 76.91\\
  \hline
  WordSup~\cite{2017wordsup} 		& * & * & * & 87.53 & 93.34 & 90.34 & 77.03 & 79.33 & 78.16\\
  \hline
  RRPN~\cite{ma2017arbitrary} 		& * & * & * & 88 & \bf{95} & 91 & 77.13 & 83.52 & 80.20\\
  \hline
  EAST~\cite{2017east} 			& * & * & * & * & * & * & 78.33 & 83.27 & 80.72\\
  \hline
  He et al.~\cite{he2017direct} 			& 81 & 92 & 86 & * & * & * & 82 & 80 & 81\\
  \hline
  R2CNN~\cite{jiang2017r2cnn} 		& * & * & * & 82.59 & 93.55 & 87.73 & 79.68 & 85.62 & 82.54\\
  \hline
  FTSN~\cite{dai2017fused} 			& 81 & 92 & 86 & * & * & * & 80.07 & 88.65 & 84.14\\
  \hline
  Li et al.~\cite{2017towards} 	 & 80.5 & 91.4 & 85.6 & * & * & *  & * & * & * \\
  \hline\hline
  TextNet-detection only					& 82.01 & 83.40 & 84.62 & 82.15 &87.40 & 84.70 & 80.83 & 85.79 & 83.24\\
  \hline 
  TextNet 	   					& \bf{89.39} & \bf{93.26} & \bf{91.28} & \bf{89.19} & 93.62 & \bf{91.35}  & \bf{85.41} & \bf{89.42} & \bf{87.37}\\
  \hline
\end{tabular}
\end{center}
\end{table}
\vspace{-2em}
\begin{table}
\begin{center}
\footnotesize 
\caption{\footnotesize End-to-end text reading and word spotting results on ICDAR-13. Note that S, W and G are short for strong, weakly and generic conditions, and TextNet indicates the multi-scale testing of the proposed model.}
\label{tab:icdar-13-end2end}
\begin{tabular}{|c||c|c|c||c|c|c|}
  \hline
  \multirow{2}{*}{Method} &
  \multicolumn{3}{c||}{End-to-end} &
  \multicolumn{3}{c|}{Word spotting} \\ \cline{2-7} 
   & S & W & G & S & W & G \\
  \hline\hline
  Deep2Text II+ & 81.81 & 79.49 & 76.99 & 84.84 & 83.43 & 78.90 \\
  \hline
  TextBoxes~\cite{liao2017textboxes} & \bf{91.57} & 89.65 & 83.89 & 93.90 & 91.95 & 85.92 \\
  \hline
  Li et al.~\cite{2017towards} & 91.08 & \bf{89.81} & \bf{84.59} & 94.16 & 92.42 & \bf{88.20} \\
  \hline
  TextSpotter~\cite{2017deep_textspotter} & 89.0 & 86.0 & 77.0 & 92.0 & 89.0 & 81.0 \\
  \hline
  \hline 
  TextNet 	   & 89.77 & 88.80 &  82.96 & \bf{94.59} & \bf{93.48} & 86.99 \\
  \hline
\end{tabular}
\end{center}
\vspace{-3.5em}
\end{table}
\begin{table}
\begin{center}
\footnotesize 
\caption{\footnotesize End-to-end text reading and word spotting results on ICDAR-15 for comparison. Note that * indicates that the corresponding results have not been reported in the original paper.}
\label{tab:icdar-15-end2end}
\begin{tabular}{|c||c|c|c||c|c|c|}
  \hline
  \multirow{2}{*}{Method} &
  \multicolumn{3}{c||}{End-to-end} &
  \multicolumn{3}{c|}{Word spotting} \\ \cline{2-7} 
  & S & W & G & S & W & G \\
  \hline\hline
  HUST MCLAB \cite{shi2017end} & 67.86 & * & * & 70.57 & * & * \\
  \hline
  TextProposals+DictNet & 53.30 & 49.61 & 47.18 & 56.00 & 52.26 & 49.73 \\
  \hline
  TextSpotter~\cite{2017deep_textspotter} & 54.0 & 51.0 & 47.0 & 58.0 & 53.0 & 51.0 \\
  \hline
  \hline 
  TextNet        & \bf{78.66}  &  \bf{74.90}  &  \bf{60.45}  & \bf{82.38}  &  \bf{78.43}  & \bf{62.36}\\
  \hline
\end{tabular}
\end{center}
\vspace{-1.5em}
\end{table}
\vspace{-1em}
\subsubsection{End-to-end results} We conduct experiments on ICDAR-13, ICDAR-15, and Total-Text datasets, evaluating the results on regular and irregular benchmarks.
For fair comparisons, the quantitative results of TextNet on ICDAR-13 and ICDAR-15 are shown in Tab.~\ref{tab:icdar-13-end2end} and Tab.~\ref{tab:icdar-15-end2end} in terms of F-measure for end-to-end and word spotting tasks, following the evaluation protocols in ICDAR competitions under strong, weakly and generic conditions.
These results demonstrate its superior performance over state-of-the-art approaches on ICDAR benchmarks, especially on ICDAR-15 dataset. On ICDAR-13 dataset, the end-to-end trainable model proposed by Li et al. \cite{2017towards} shows better performance for horizontal cases, but fails to cover cases in multi-orientations on ICDAR-15 dataset. For irregular cases, experimental results on Total-Text is illustrated in Tab.~\ref{tab:total-text-det}. We report the text localization and end-to-end results in terms of recall, precision and F-measure. In the end-to-end task, the score of F-measure on Total-text is to evaluate the raw model output according to ground-truth labels without any vocabularies and language models.  The results on irregular text demonstrate that the proposed TextNet have shown dramatic improvement over the baseline algorithm, which validates the effectiveness of the attention mechanism to tackle irregular text, especially for curve cases. Note that the most recent approaches \cite{2018fots} are designed for regular text and are difficult to cover curve cases.
\vspace{-1em}
\subsection{Qualitative Results}
\begin{figure}
\centering
\SetFigLayout{3}{2}
  \subfigure{\includegraphics[width=0.32\textwidth, height=0.23\textwidth]{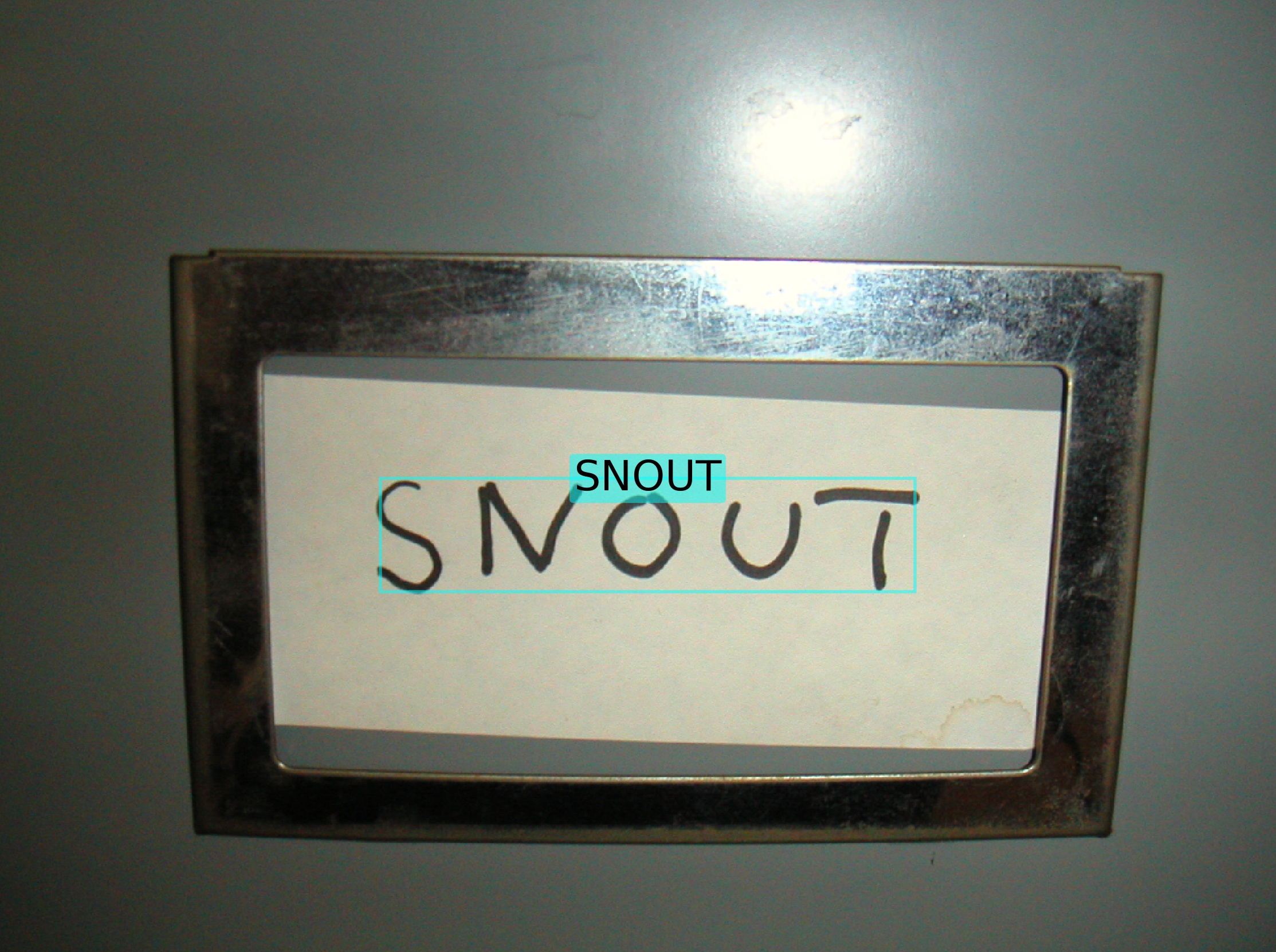}}
  \subfigure{\includegraphics[width=0.32\textwidth, height=0.23\textwidth]{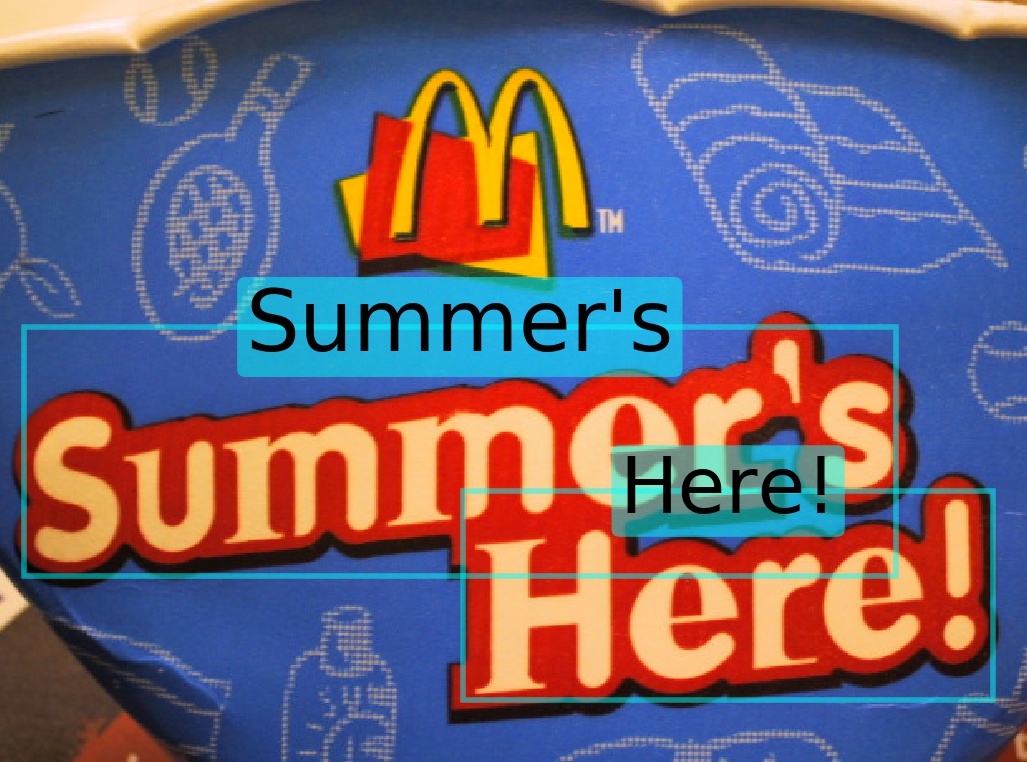}}
  \subfigure{\includegraphics[width=0.32\textwidth, height=0.23\textwidth]{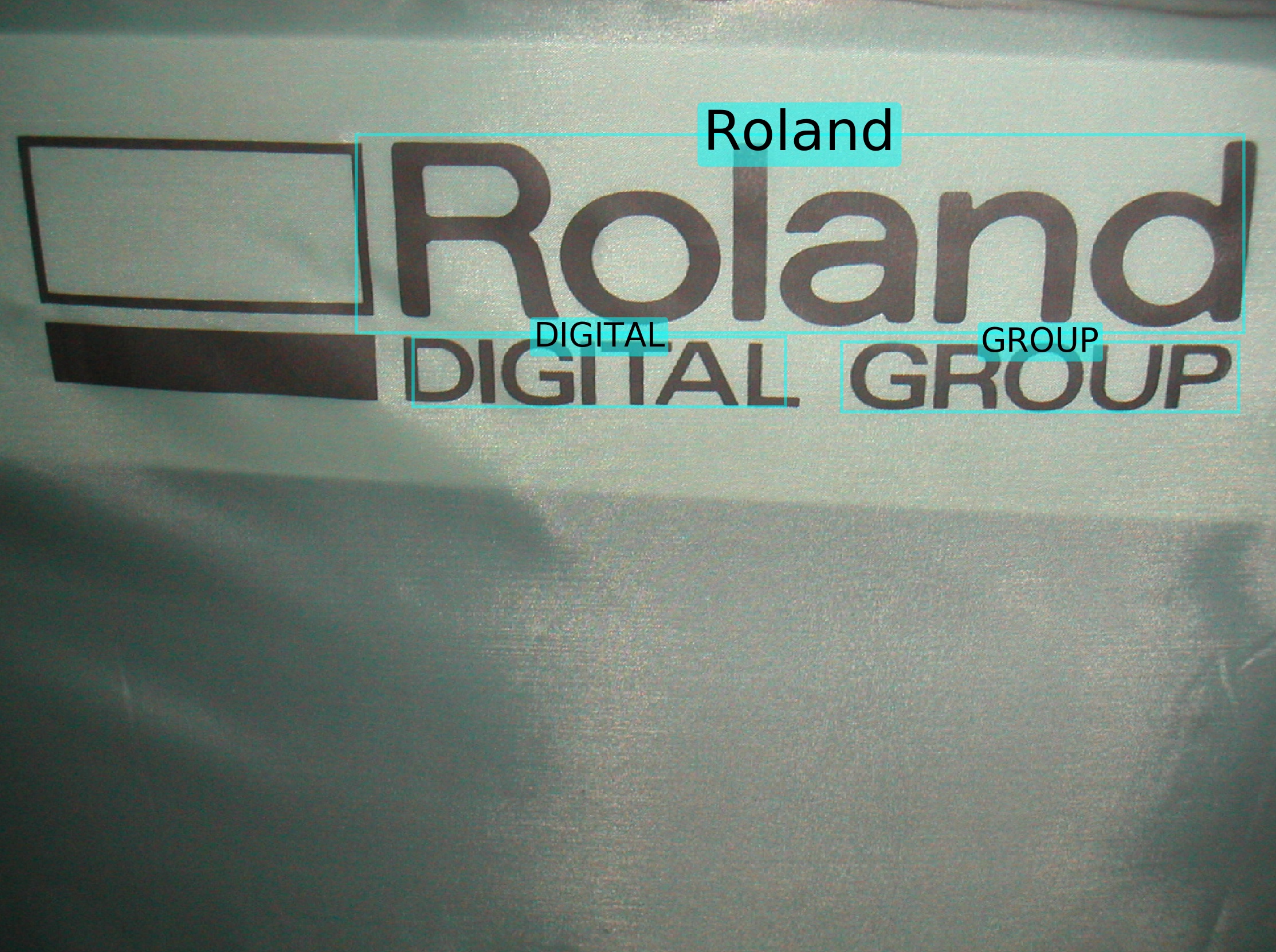}}\vspace{-0.5em}
  \subfigure{\includegraphics[width=0.32\textwidth, height=0.23\textwidth]{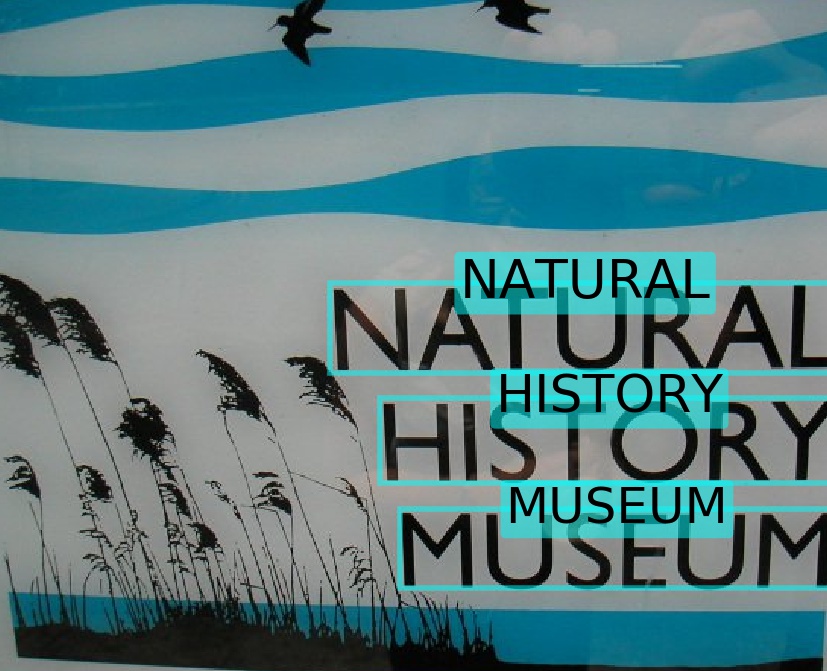}}
  \subfigure{\includegraphics[width=0.32\textwidth, height=0.23\textwidth]{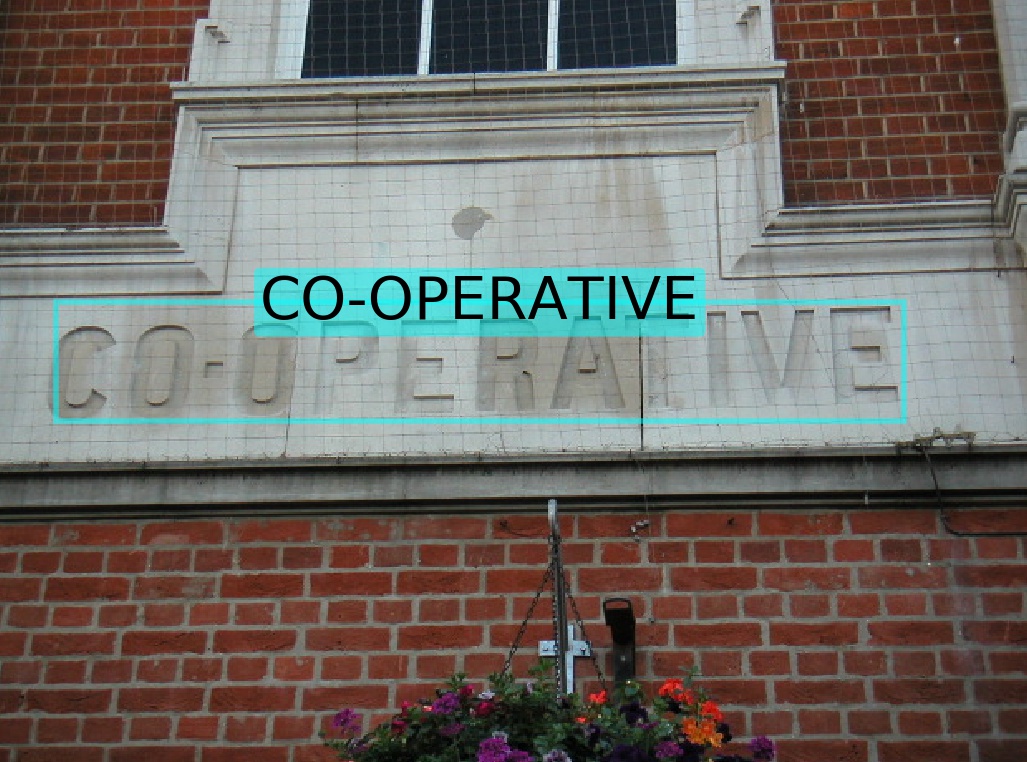}}
  \subfigure{\includegraphics[width=0.32\textwidth, height=0.23\textwidth]{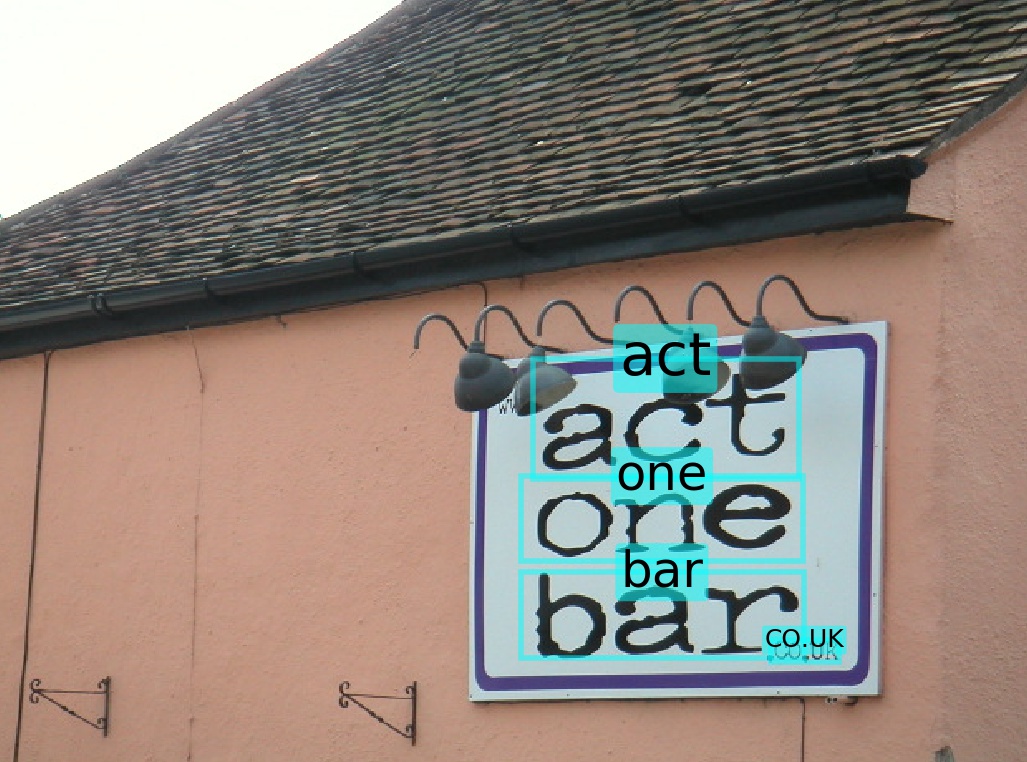}}\vspace{-1em}
\caption{\footnotesize End-to-end visualized results of TextNet on ICDAR-13 dataset.}
\label{fig:visual_results-icdar13}
\vspace{-1.5em}
\end{figure}

\begin{figure}
\centering
\SetFigLayout{3}{2}
  \subfigure
  {\includegraphics[width=0.49\textwidth, height=0.26\textwidth]{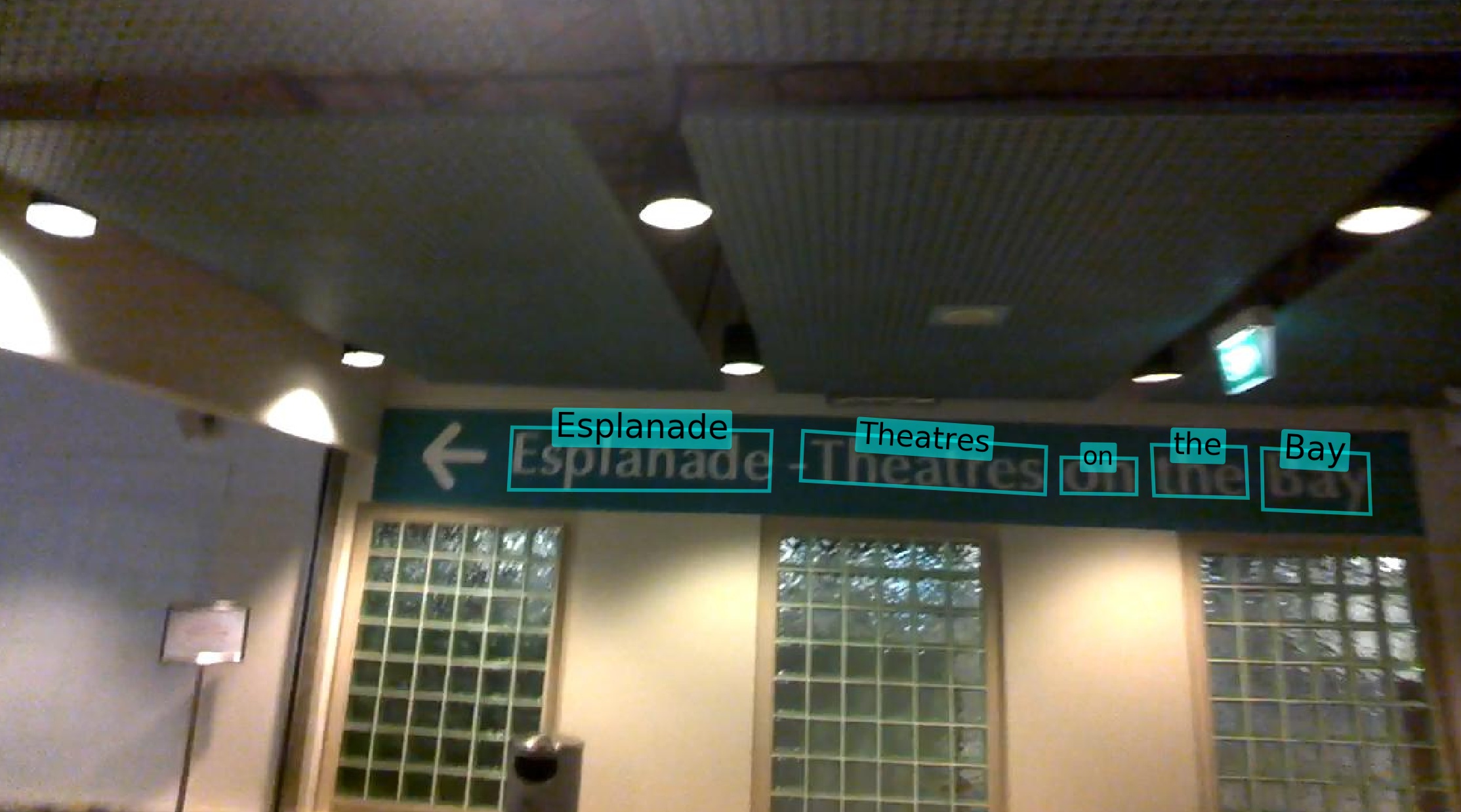}}
  \subfigure
  {\includegraphics[width=0.49\textwidth, height=0.26\textwidth]{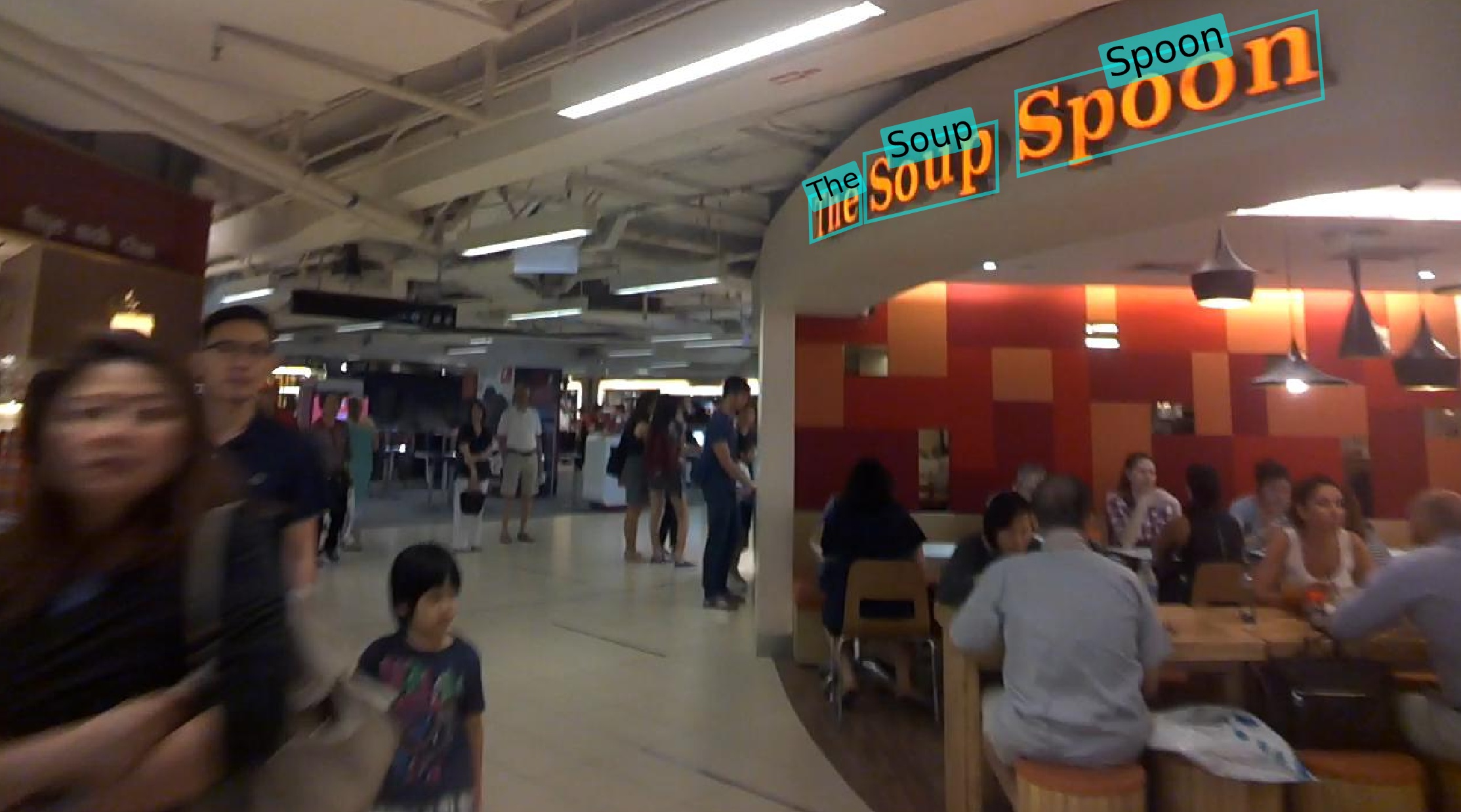}}\vspace{-1em}\\
  \subfigure
  {\includegraphics[width=0.49\textwidth, height=0.26\textwidth]{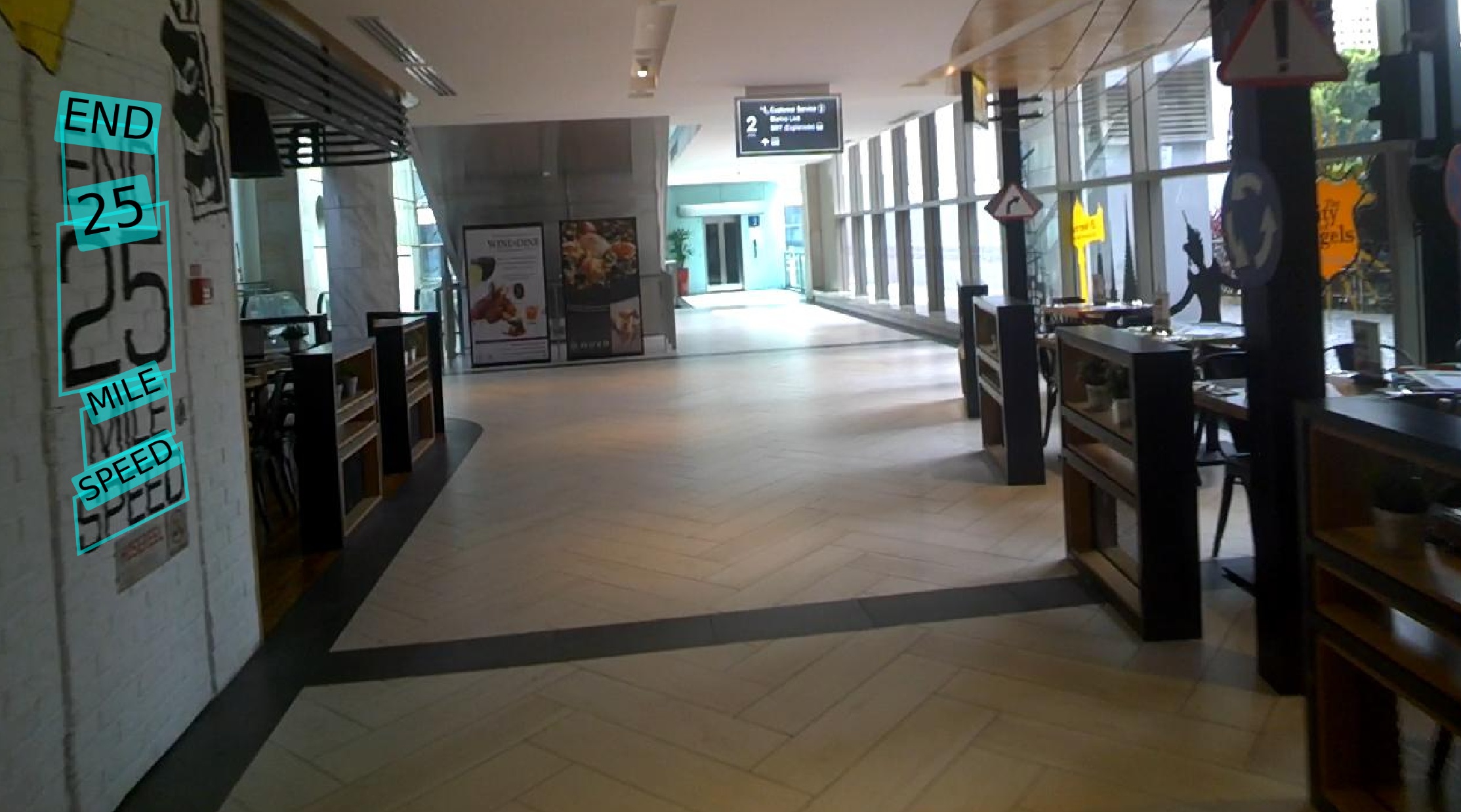}}
  \subfigure
  {\includegraphics[width=0.49\textwidth, height=0.26\textwidth]{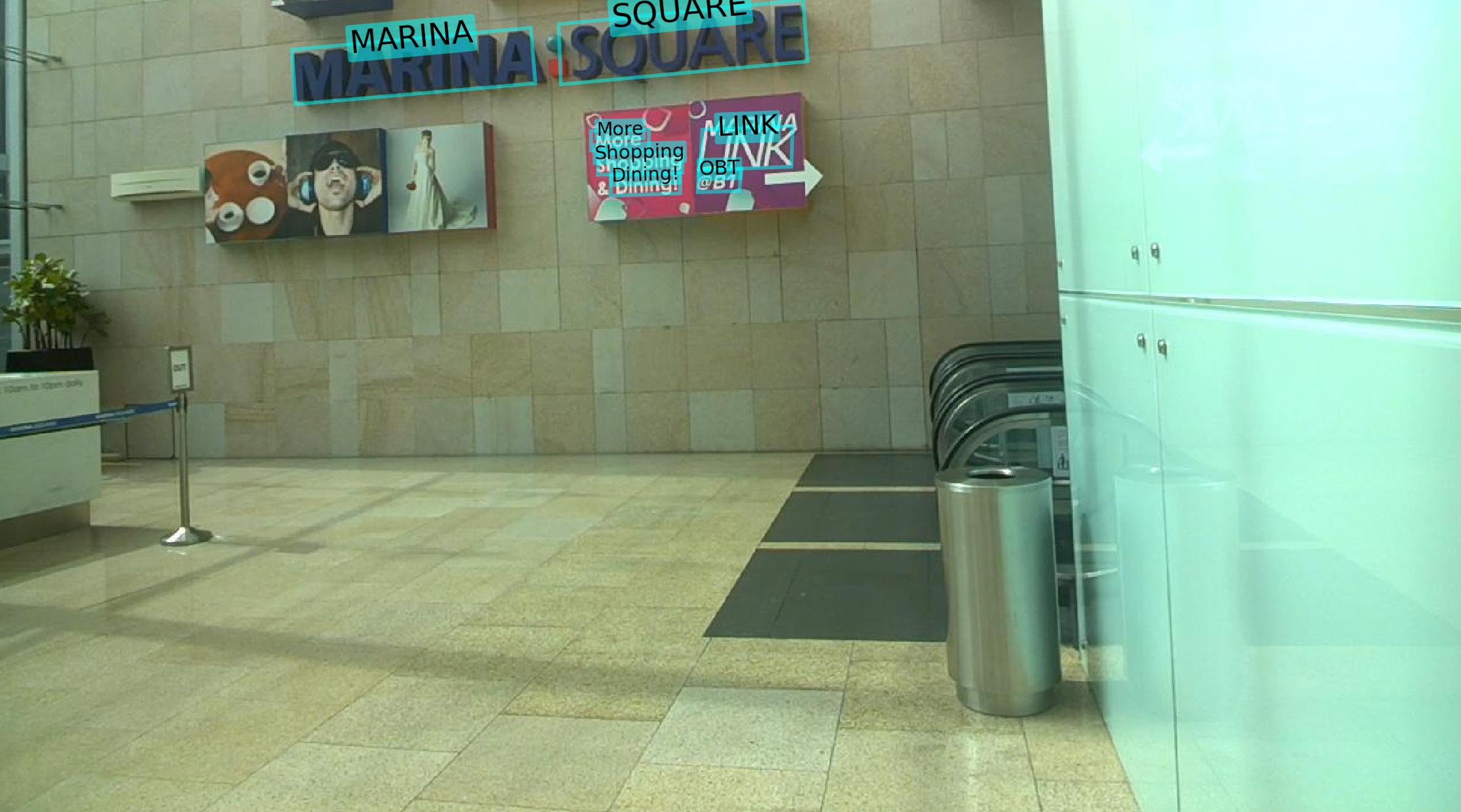}}
  \vspace{-1em}\\
   \subfigure
  {\includegraphics[width=0.49\textwidth, height=0.26\textwidth]{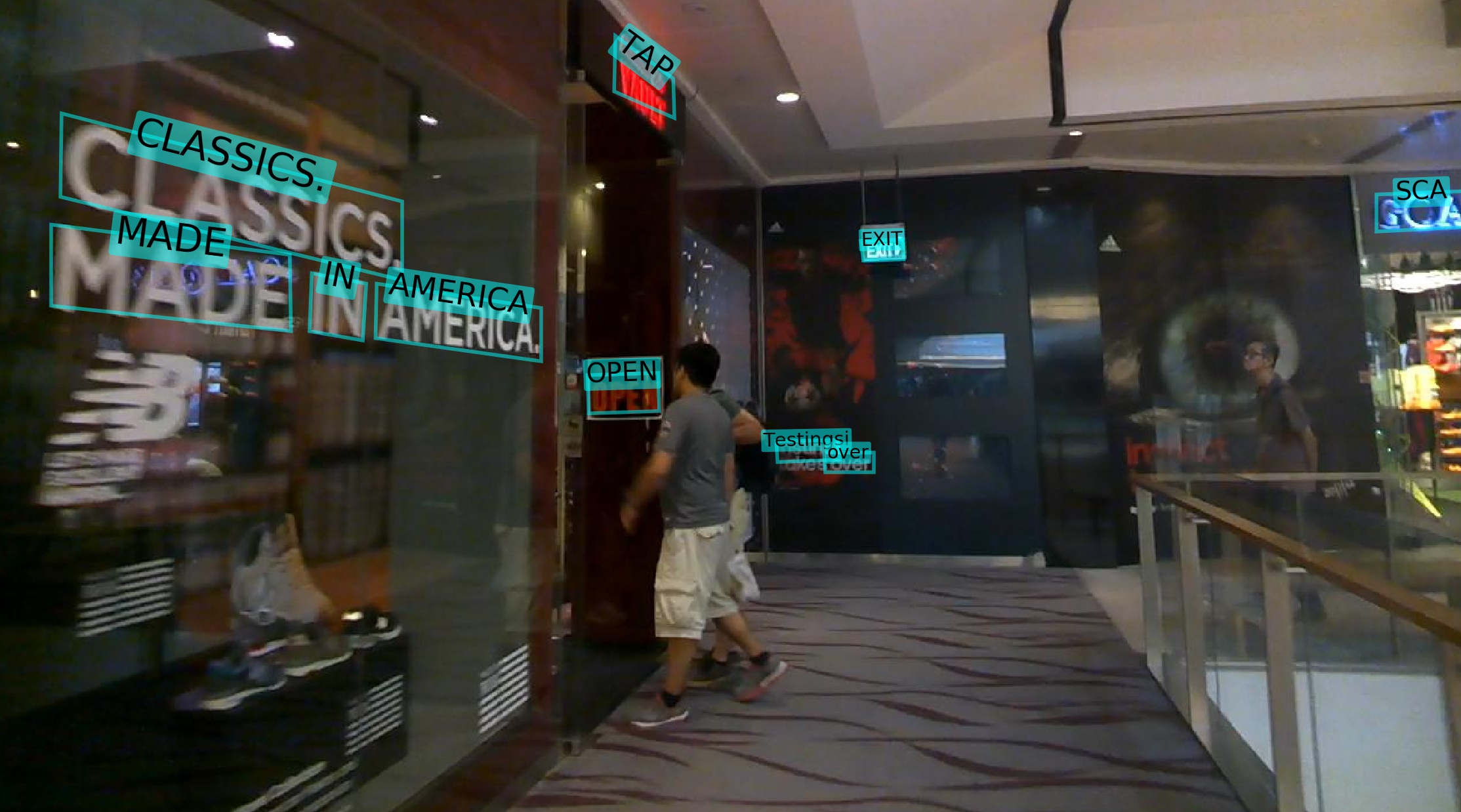}}
  \subfigure
  {\includegraphics[width=0.49\textwidth, height=0.26\textwidth]{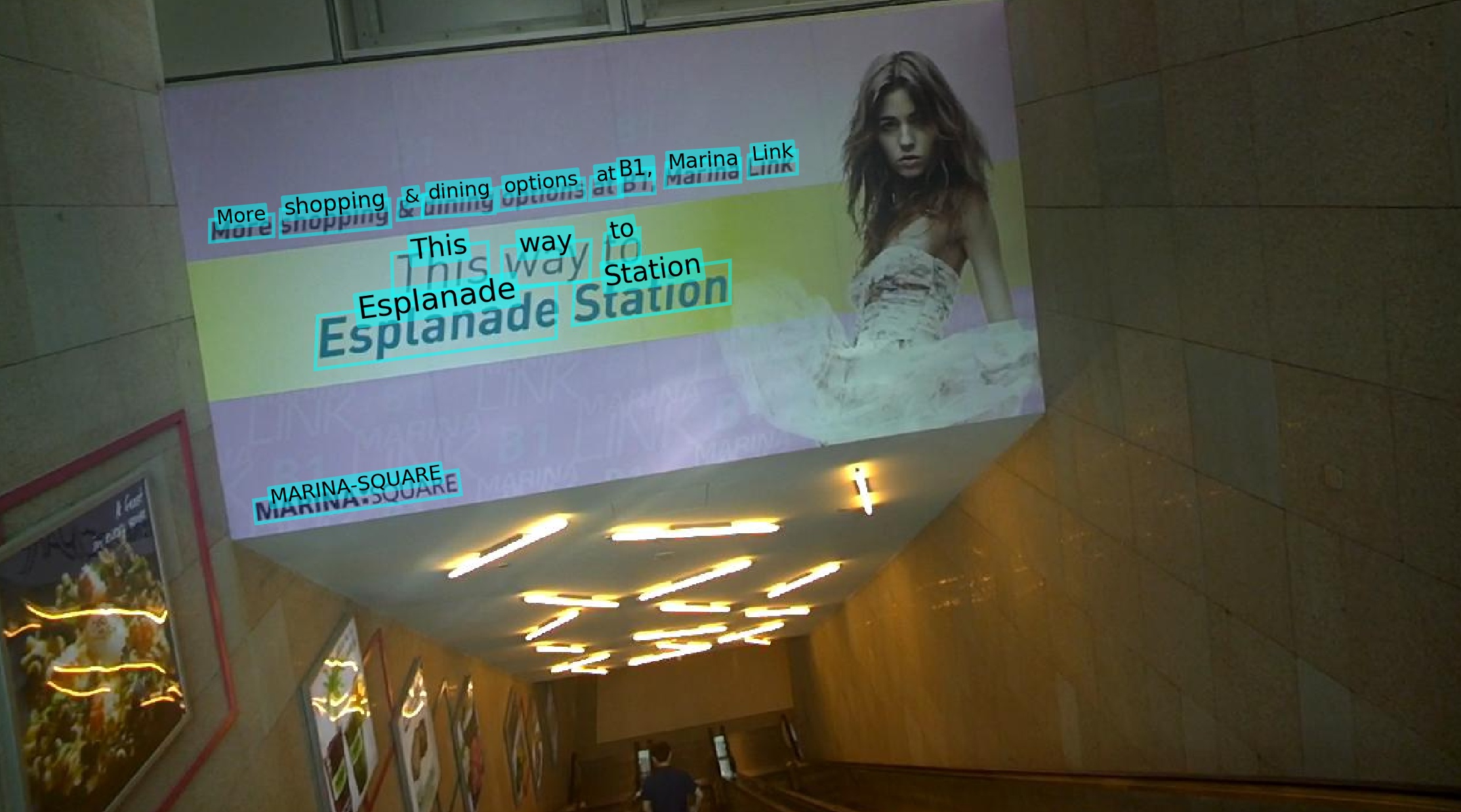}}
  \vspace{-1em}
\caption{\footnotesize End-to-end visualized results of TextNet on ICDAR-15 dataset.}
\label{fig:visual_results-icdar15}
\end{figure}

\begin{figure}
\centering
\SetFigLayout{2}{3}
  \subfigure
  {\includegraphics[width=0.325\textwidth, height=0.18\textwidth]{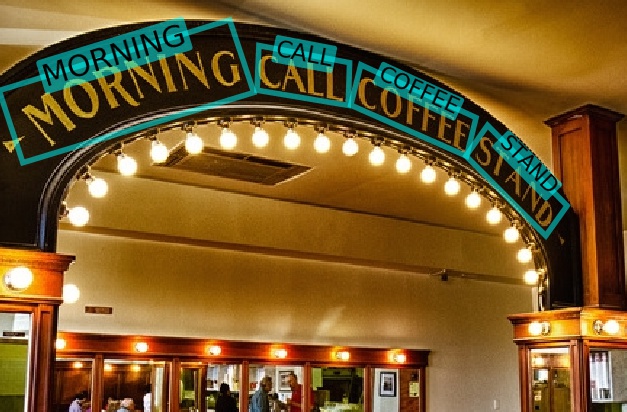}}
  \subfigure
  {\includegraphics[width=0.325\textwidth, height=0.18\textwidth]{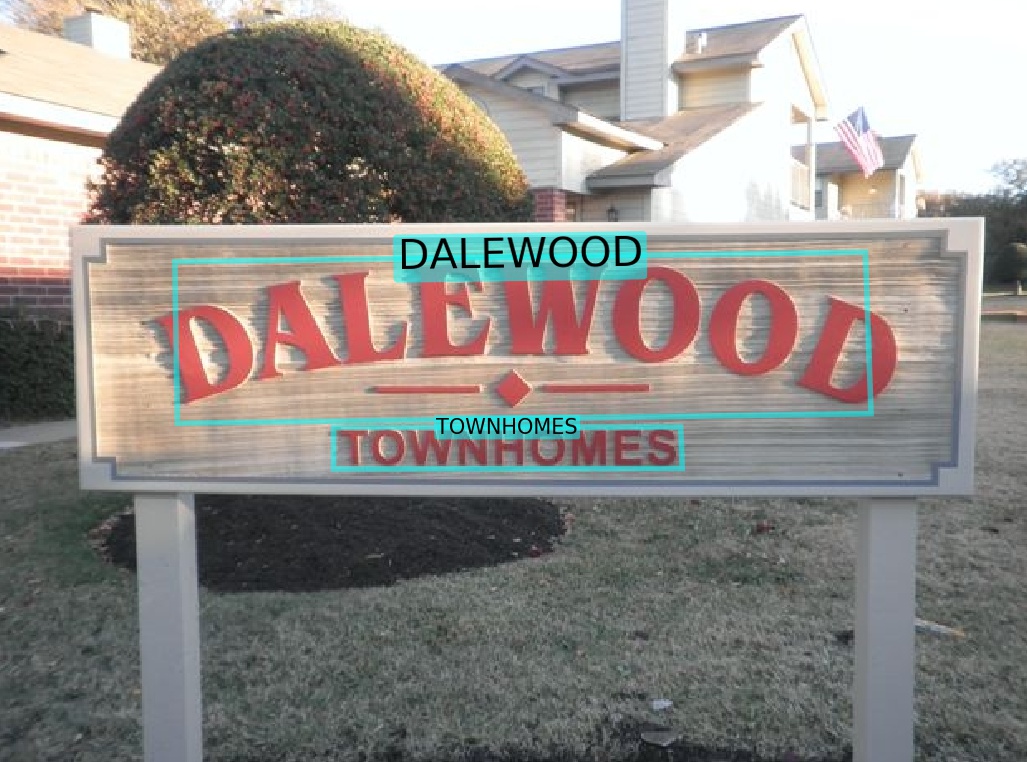}}
  \subfigure
  {\includegraphics[width=0.325\textwidth, height=0.18\textwidth]{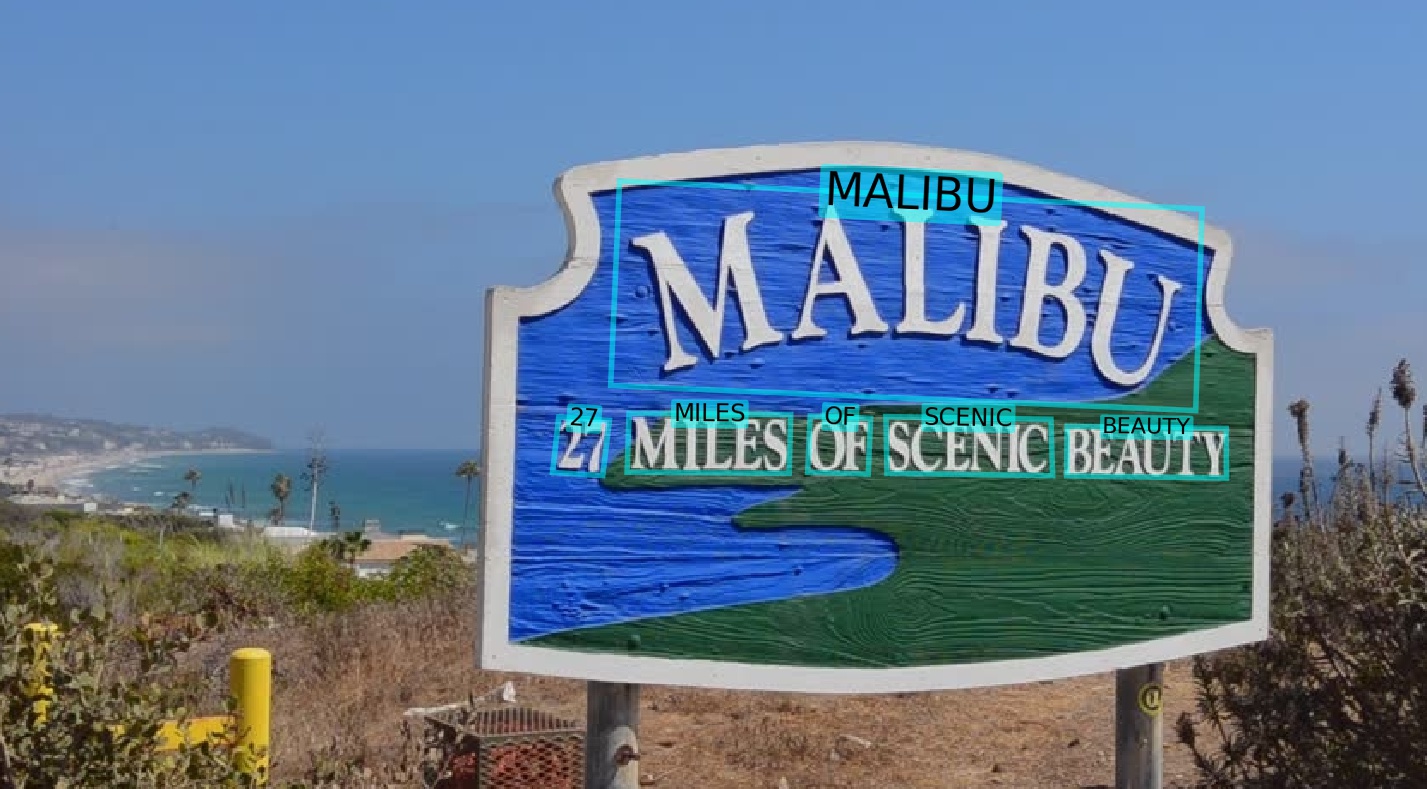}}\vspace{-1em}\\
  \subfigure
  {\includegraphics[width=0.325\textwidth, height=0.18\textwidth]{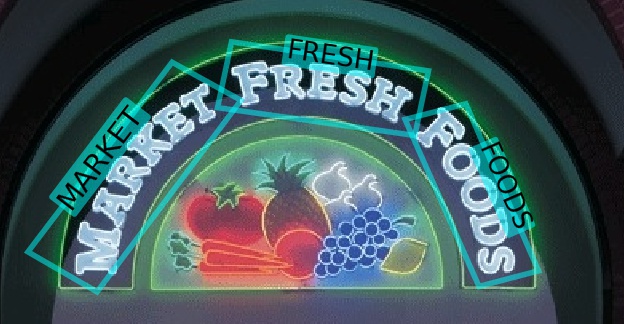}}   
  \subfigure
  {\includegraphics[width=0.325\textwidth, height=0.18\textwidth]{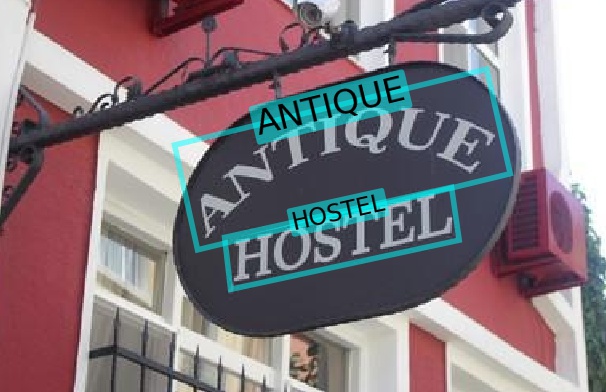}}
   \subfigure
  {\includegraphics[width=0.325\textwidth, height=0.18\textwidth]{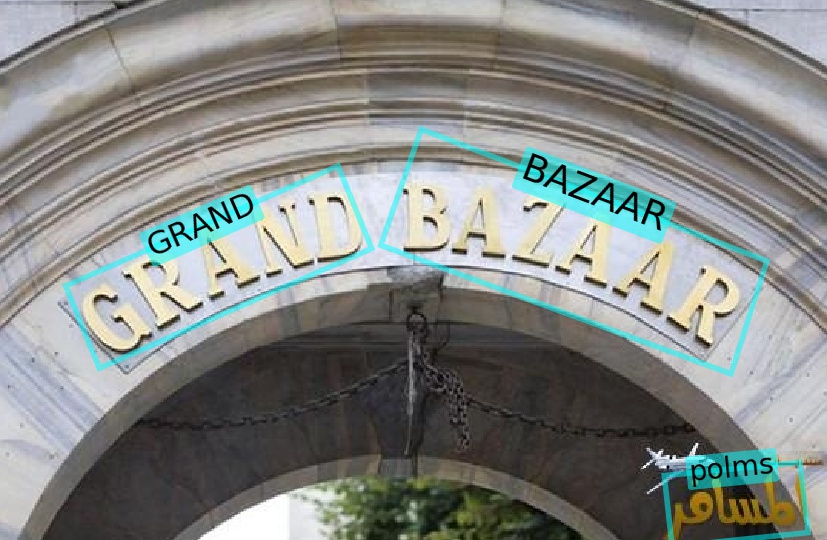}}\vspace{-1em}\\
\caption{\footnotesize End-to-end visualized results of TextNet on Total-Text dataset.}
\label{fig:visual_results-tt}
\end{figure}

Qualitative results of TextNet on ICDAR-13, ICDAR-15 and Total-Text datasets are shown in Fig.~\ref{fig:visual_results-icdar13}, Fig.~\ref{fig:visual_results-icdar15}, and Fig.~\ref{fig:visual_results-tt}, respectively.  The localization quadrangles and the corresponding predicted text are drawn in figures. From these visual results, we can see that the proposed approach is able to tackle regular and irregular cases. The proposed algorithm can accurately localize and recognize these samples in an end-to-end fashion, which validates its effectiveness of the proposed approach. 

\begin{table}
\begin{center}
\footnotesize 
\caption{\footnotesize Text detection and end-to-end recognition performance on irregular dataset, i.e., Total-Text. The detection results exactly follows the evaluation rules of Total-Text~\cite{chng17tt}. Note that * indicates that the corresponding results have not been reported yet in the original paper.}
\label{tab:total-text-det}
\begin{tabular}{|c||c|c|c||c|c|c|}
  \hline
  \multirow{2}{*}{Method} &
  \multicolumn{3}{c||}{Detection} &
  \multicolumn{3}{c|}{End-to-end} \\ \cline{2-7} 
  & Recall & Precision & F-measure & Recall & Precision & F-measure\\
  \hline\hline
  DeconvNet \cite{chng17tt} & 33 & 40 & 36 & * & * & *\\
  \hline
  \hline 
  Two-stage approach     	& \bf{59.80}  & 61.05 &  60.42  & 43.10  & 47.12  & 45.02 \\
  \hline
  TextNet   	& 59.45  & \bf{68.21} &  \bf{63.53}  & \bf{56.39}  & \bf{51.85}  & \bf{54.02} \\
  \hline
\end{tabular}
\end{center}
\vspace{-1em}
\end{table}

\vspace{-1em}
\subsection{Speed and Model Size}
To evaluate the speed of TextNet, we calculate the average time cost during the testing stage. On the ICDAR-13 dataset, we can achieve $370.6$ ms in terms of average time cost using ResNet-50 without any model compression and acceleration. Note that the long-side length of testing images is normalized to $920$, and the time cost is evaluated using a single Tesla P40 GPU.
The total number of parameters of TextNet is $30$M including ResNet-50, which includes $23$M coefficients taking the most of parameters in the proposed model. By sharing backbone network, the jointly trained model not only reduces the time cost during predicting stage but also saves almost half of parameters compared with separately trained models.

 \vspace{-1em}

\vspace{-0.5em}
\section{Conclusions}
In this paper, we have proposed a fully end-to-end trainable network, i.e., \textit{TextNet}, which is capable of simultaneously localizing and recognizing regular and irregular text. The proposed network can extract multi-scale image features by scale-aware attention mechanism, and generate word proposals by direct regression of quadrangles to cover regular and irregular text regions. To further extract features-of-interest from text proposals, we have proposed to encode the well aligned feature maps in RNN, and utilize the spatial attention mechanism to generate text sequences. The experimental results on benchmark datasets have shown that the proposed \textit{TextNet} can achieve superior performance on ICDAR regular datasets, and outperform existing approaches for irregular cases by a large margin.

%
%
%
%
\bibliographystyle{splncs04}
\bibliography{egbib}

\begin{thebibliography}{10}
\providecommand{\url}[1]{\texttt{#1}}
\providecommand{\urlprefix}{URL }
\providecommand{\doi}[1]{https://doi.org/#1}

\bibitem{bahdanau2014neural}
Bahdanau, D., Cho, K., Bengio, Y.: Neural machine translation by jointly
  learning to align and translate. arXiv preprint arXiv:1409.0473  (2014)

\bibitem{bissacco2013photoocr}
Bissacco, A., Cummins, M., Netzer, Y., Neven, H.: Photo-{OCR}: Reading text in
  uncontrolled conditions. In: Proc. of ICCV. pp. 785--792 (2013)

\bibitem{bluche2016joint}
Bluche, T.: Joint line segmentation and transcription for end-to-end
  handwritten paragraph recognition. In: Proc. of NIPS. pp. 838--846 (2016)

\bibitem{bluche2016scan}
Bluche, T., Louradour, J., Messina, R.: Scan, attend and read: End-to-end
  handwritten paragraph recognition with mdlstm attention. arXiv preprint
  arXiv:1604.03286  (2016)

\bibitem{2017deep_textspotter}
Bu{\v{s}}ta, M., Neumann, L., Matas, J.: Deep textspotter: An end-to-end
  trainable scene text localization and recognition framework. Proc. of ICCV
  (2017)

\bibitem{chen2016attention}
Chen, L.C., Yang, Y., Wang, J., Xu, W., Yuille, A.L.: Attention to scale:
  Scale-aware semantic image segmentation. In: Proc. of CVPR. pp. 3640--3649
  (2016)

\bibitem{chng17tt}
Chng, C.K., Chan, C.S.: Total-text: {A} comprehensive dataset for scene text
  detection and recognition. In: Proc. of ICDAR (2017)

\bibitem{dai2017fused}
Dai, Y., Huang, Z., Gao, Y., Chen, K.: Fused text segmentation networks for
  multi-oriented scene text detection. arXiv preprint arXiv:1709.03272  (2017)

\bibitem{girshick2015fast}
Girshick, R.: Fast {R-CNN}. arXiv preprint arXiv:1504.08083  (2015)

\bibitem{graves2006connectionist}
Graves, A., Fern{\'a}ndez, S., Gomez, F., Schmidhuber, J.: Connectionist
  temporal classification: labelling unsegmented sequence data with recurrent
  neural networks. In: Proc. of ICML. pp. 369--376. ACM (2006)

\bibitem{graves2013speech}
Graves, A., Mohamed, A.r., Hinton, G.: Speech recognition with deep recurrent
  neural networks. In: Proc. of ICASSP. pp. 6645--6649 (2013)

\bibitem{gupta2016synthetic}
Gupta, A., Vedaldi, A., Zisserman, A.: Synthetic data for text localisation in
  natural images. In: Proc. of CVPR. pp. 2315--2324 (2016)

\bibitem{he2017mask}
He, K., Gkioxari, G., Doll{\'a}r, P., Girshick, R.: Mask r-cnn. In: Proc. of
  ICCV. pp. 2980--2988 (2017)

\bibitem{he2016deep}
He, K., Zhang, X., Ren, S., Sun, J.: Deep residual learning for image
  recognition. In: Proc. of CVPR. pp. 770--778 (2016)

\bibitem{he2016reading}
He, P., Huang, W., Qiao, Y., Loy, C.C., Tang, X.: Reading scene text in deep
  convolutional sequences. In: Proc. of AAAI. vol.~16, pp. 3501--3508 (2016)

\bibitem{2018textspotter}
He, T., Tian, Z., Huang, W., Shen, C., Qiao, Y., Sun, C.: An end-to-end
  textspotter with explicit alignment and attention. CoRR
  \textbf{abs/1803.03474} (2018)

\bibitem{he2017direct}
He, W., Zhang, X.Y., Yin, F., Liu, C.L.: Deep direct regression for
  multi-oriented scene text detection. In: Proc. of ICCV (2017)

\bibitem{2017wordsup}
Hu, H., Zhang, C., Luo, Y., Wang, Y., Han, J., Ding, E.: Wordsup: Exploiting
  word annotations for character based text detection. In: Proc. of ICCV (2017)

\bibitem{huang2015densebox}
Huang, L., Yang, Y., Deng, Y., Yu, Y.: Densebox: Unifying landmark localization
  with end to end object detection. arXiv preprint arXiv:1509.04874  (2015)

\bibitem{jaderberg2016reading}
Jaderberg, M., Simonyan, K., Vedaldi, A., Zisserman, A.: Reading text in the
  wild with convolutional neural networks. International Journal of Computer
  Vision  \textbf{116}(1),  1--20 (2016)

\bibitem{jaderberg2015spatial}
Jaderberg, M., Simonyan, K., Zisserman, A., et~al.: Spatial transformer
  networks. In: Proc. of NIPS. pp. 2017--2025 (2015)

\bibitem{jiang2017r2cnn}
Jiang, Y., Zhu, X., Wang, X., Yang, S., Li, W., Wang, H., Fu, P., Luo, Z.:
  R2cnn: rotational region cnn for orientation robust scene text detection.
  arXiv preprint arXiv:1706.09579  (2017)

\bibitem{karatzas15icdar}
Karatzas, D., et~al: {ICDAR} 2015 competition on robust reading. In: Proc. of
  ICDAR. pp. 1156--1160. IEEE (2015)

\bibitem{lee2016recursive}
Lee, C.Y., Osindero, S.: Recursive recurrent nets with attention modeling for
  {ocr} in the wild. In: Proc. of CVPR. pp. 2231--2239 (2016)

\bibitem{2017towards}
Li, H., Wang, P., Shen, C.: Towards end-to-end text spotting with convolutional
  recurrent neural networks. Proc. of ICCV  (2017)

\bibitem{liao2018textboxes++}
Liao, M., Shi, B., Bai, X.: Textboxes++: A single-shot oriented scene text
  detector. arXiv preprint arXiv:1801.02765  (2018)

\bibitem{liao2017textboxes}
Liao, M., Shi, B., Bai, X., Wang, X., Liu, W.: Text{B}oxes: A fast text
  detector with a single deep neural network. In: Proc. of AAAI. pp. 4161--4167
  (2017)

\bibitem{lin2017feature}
Lin, T.Y., Doll{\'a}r, P., Girshick, R., He, K., Hariharan, B., Belongie, S.:
  Feature pyramid networks for object detection. In: Proc. of CVPR (2017)

\bibitem{liu2016ssd}
Liu, W., Anguelov, D., Erhan, D., Szegedy, C., Reed, S., Fu, C.Y., Berg, A.C.:
  Ssd: Single shot multibox detector. In: Proc. of ECCV. pp. 21--37. Springer
  (2016)

\bibitem{2018fots}
Liu, X., Liang, D., Yan, S., Chen, D., Qiao, Y., Yan, J.: Fots: Fast oriented
  text spotting with a unified network. arXiv preprint arXiv:1801.01671  (2018)

\bibitem{2017deep_matching}
Liu, Y., Jin, L.: Deep matching prior network: Toward tighter multi-oriented
  text detection. In: Proc. of CVPR (2017)

\bibitem{ma2017arbitrary}
Ma, J., Shao, W., Ye, H., Wang, L., Wang, H., Zheng, Y., Xue, X.:
  Arbitrary-oriented scene text detection via rotation proposals. arXiv
  preprint arXiv:1703.01086  (2017)

\bibitem{neumann12mser}
Neumann, L., Matas, J.: Real-time scene text localization and recognition. In:
  Proc. of CVPR. pp. 3538--3545 (2012)

\bibitem{redmon2016yolo9000}
Redmon, J., Farhadi, A.: Yolo9000: better, faster, stronger. arXiv preprint
  \textbf{1612} (2016)

\bibitem{ren2015faster}
Ren, S., He, K., Girshick, R., Sun, J.: Faster {R}-{CNN}: Towards real-time
  object detection with region proposal networks. In: Proc. of NIPS. pp. 91--99
  (2015)

\bibitem{ronneberger2015u}
Ronneberger, O., Fischer, P., Brox, T.: U-net: Convolutional networks for
  biomedical image segmentation. In: International Conference on Medical image
  computing and computer-assisted intervention. pp. 234--241. Springer (2015)

\bibitem{shi2017detecting}
Shi, B., Bai, X., Belongie, S.: Detecting oriented text in natural images by
  linking segments. Proc. of CVPR  (2017)

\bibitem{shi2017end}
Shi, B., Bai, X., Yao, C.: An end-to-end trainable neural network for
  image-based sequence recognition and its application to scene text
  recognition. IEEE Transactions on Pattern Analysis and Machine Intelligence
  \textbf{39}(11),  2298--2304 (2017)

\bibitem{shi2016robust}
Shi, B., Wang, X., Lyu, P., Yao, C., Bai, X.: Robust scene text recognition
  with automatic rectification. In: Proc. of CVPR. pp. 4168--4176 (2016)

\bibitem{tian2016ctpn}
Tian, Z., Huang, W., He, T., He, P., Qiao, Y.: Detecting text in natural image
  with connectionist text proposal network. In: Proc. of ECCV. pp. 56--72
  (2016)

\bibitem{wang2012end}
Wang, T., Wu, D.J., Coates, A., Ng, A.Y.: End-to-end text recognition with
  convolutional neural networks. In: Proc. of ICPR. pp. 3304--3308 (2012)

\bibitem{2017attention}
Wojna, Z., Gorban, A., Lee, D.S., Murphy, K., Yu, Q., Li, Y., Ibarz, J.:
  Attention-based extraction of structured information from street view
  imagery. Proc. of ICDAR  (2017)

\bibitem{xu2015show}
Xu, K., Ba, J., Kiros, R., Cho, K., Courville, A., Salakhudinov, R., Zemel, R.,
  Bengio, Y.: Show, attend and tell: Neural image caption generation with
  visual attention. In: Proc. of ICML. pp. 2048--2057 (2015)

\bibitem{yang2017learning}
Yang, X., He, D., Zhou, Z., Kifer, D., Giles, C.L.: Learning to read irregular
  text with attention mechanisms. In: Proc. of IJCAI (2017)

\bibitem{2017east}
Zhou, X., Yao, C., Wen, H., Wang, Y., Zhou, S., He, W., Liang, J.: East: An
  efficient and accurate scene text detector. In: Proc. of CVPR (2017)

\end{thebibliography}

\end{document}